\pdfoutput=1

\documentclass[11pt]{article}

\usepackage[final]{acl}

\usepackage{times}
\usepackage{latexsym}
\usepackage{csquotes}
\usepackage{booktabs}
\usepackage{makecell}
\usepackage{multirow} 
\usepackage{array}
\newcolumntype{M}[1]{>{\centering\arraybackslash}m{#1}}
\usepackage[T1]{fontenc}

\usepackage[utf8]{inputenc}

\usepackage{microtype}

\usepackage{inconsolata}

\usepackage{graphicx}
\usepackage{algorithm}
\usepackage{algpseudocode}
\usepackage{amsmath} 

\usepackage[most]{tcolorbox}
\usepackage{xcolor}
\usepackage{enumitem}
\usepackage{amssymb}

\usepackage{colortbl} 
\PassOptionsToPackage{table}{xcolor}
\PassOptionsToPackage{dvipsnames}{xcolor}

\definecolor{hj_blue}{HTML}{B3C8CF}
\definecolor{hj_gray}{HTML}{BFCFE7}
\definecolor{hj_deep}{HTML}{749BC2}

\usepackage[nolist]{acronym}
\begin{acronym}
    \acro{HCI}{Human-Computer Interaction}
    \acro{LLM}{Large Language Model}
    \acro{VLMs}{Vision-Language Models}
    \acro{CS}{Computer Science}
    \acro{RAG}{Retrieval-Augmented Generation}
    \acro{AI}{Artificial Intelligent}
    \acro{MMR}{Maximal Marginal Relevance}
    \acro{VL}{Visual Language}
    \acro{MWP}{math word problem}
    \acro{AWP}{arithmetic word problem}
    \acro{VD}{visual description}
     \acro{MWPs}{math word problems}
     \acro{T2I}{Text-to-Image}
     \acro{E2V}{equation-to-visual}
     \acro{MCQs}{multiple-choice questions}
     \acro{E2V-Bench}{Equation-to-Visual Benchmark}
     \acro{DSL}{domain-specific language}
\end{acronym}
\usepackage{listings}
\usepackage{xcolor}
\usepackage[utf8]{inputenc}  

\usepackage{tikz}
\newcommand*\circled[1]{\tikz[baseline=(char.base)]{
            \node[shape=circle,draw,inner sep=1pt] (char) {#1};}}
\usepackage{todonotes}
\usepackage{marginnote}

\definecolor{TodoColor}{rgb}{1,0.7,0.6}

\definecolor{FindingsColor}{gray}{0.85}

\makeatletter\def\Hy@Warning#1{}\makeatother
\let\svthefootnote\thefootnote
\newcommand\blankfootnote[1]{%
  \let\thefootnote\relax\footnotetext{#1}%
  \let\thefootnote\svthefootnote%
}

\title{Benchmarking and Enhancing Text-to-Image Models\\ for Generating Visual Representations in Early Arithmetic Education}

\author{
    Junling Wang$^{1, 2}$ \quad
    Boqi Chen$^{1,2}$ \quad
    Heejin Do$^{1,2}$ \\
    \textbf{Mubashara Akhtar$^{1,2}$} \quad
    \textbf{April Yi Wang$^{1}$} \quad
    \textbf{
    Mrinmaya Sachan$^{1}$
    } \\ \text{} \\
  $^{1}$Department of Computer Science, ETH Zurich \quad \\
  $^2$ ETH AI Center \\
}

\begin{document}
\maketitle

\begin{abstract}

AI systems are increasingly used to support educational content creation, yet it remains unclear whether they can generate outputs that faithfully represent the pedagogical concepts they are intended to teach.
Thus, we introduce equation-to-visual generation, a task that, in contrast to conventional image generation, requires producing pedagogically meaningful visuals from arithmetic equations while precisely preserving their numerical and relational structure.
Informed by interviews with teachers and an analysis of educational materials, we construct E2V-Bench, a benchmark spanning four pedagogically grounded visual types, along with automatic metrics for evaluating visual correctness.
Our evaluation reveals that recent text-to-image (T2I) models frequently fail on this task, with errors dominated by incorrect object counts and broken relational structure.
Building on this, we explore benchmark-guided enhancement strategies. These strategies improve representative models, while the remaining gap calls for stronger numerical and relational grounding in future T2I models.

\end{abstract}

\section{Introduction}

Visuals are key pedagogical tools for teaching arithmetic concepts in early primary education. Visuals help learners transform abstract mathematical symbols into intuitive representations~\citep{cooper2018benefits}, supporting deeper understanding and problem solving skills~\citep{MAYER200285}.
For instance, to understand ``7 + 2 = 9'', learners may benefit from a visual showing seven objects together with two more to make nine 
~\citep{MAYER200285}.

\begin{figure}
    \centering
    \includegraphics[width=0.4\textwidth]{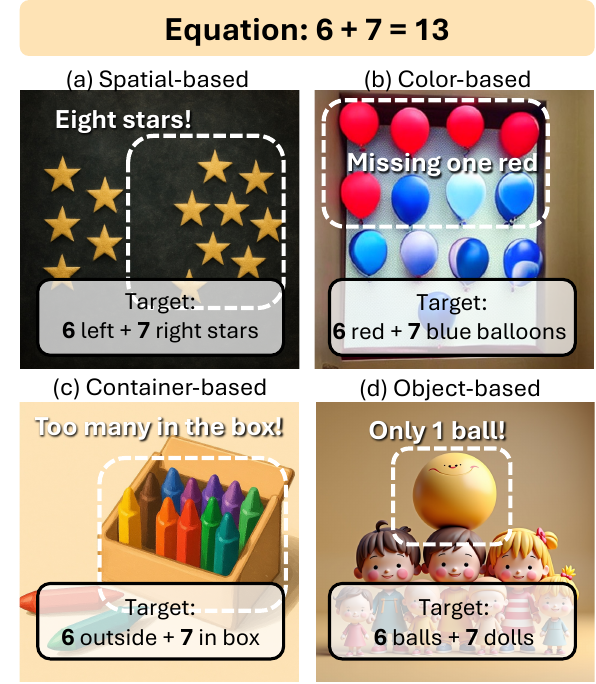}
    \caption{\textbf{Failure cases of text-to-image models.} All images are generated from the equation ``6 + 7 = 13'' using four visual types (e.g., spatial-based; see Sec.~\ref{sec:task_define}).}
    \label{fig:fig0}
    \vspace{-15pt}
\end{figure}

With the rise of multimodal tutoring systems and AI-augmented open educational resources, creating such visuals has become increasingly important, as these systems seek to provide learning materials tailored to individual learners and adapted to specific questions, solution steps, misconceptions, and backgrounds, rather than used as fixed illustrations~\citep{singh-etal-2023-enhancing,lee2025interactive}. However, manual visual creation is time-consuming and cannot support the immediate, adaptive scaffolding required in such contexts~\citep{xu2021procedural,kaitera2022developing,boonen2016s}.

This motivates automatic approaches for producing pedagogically aligned visuals on demand, with recent work beginning to explore visual generation and retrieval. Math2Visual~\citep{wang-etal-2025-generating-pedagogically} generates visuals from word problems, but relies on fixed icon sets and predefined templates, limiting its flexibility. Related works have also explored text-image matching for visual retrieval in textbook creation~\citep{singh-etal-2023-enhancing} and text-driven visual question generation~\citep{singh2019automatic, luo-etal-2024-chain}; however, these methods focus on retrieval or question construction rather than generating educational visuals.

In this work, we focus on primary school level (Grades 1-3) arithmetic education, which deals with teaching basic operations. 
We formalize the task of \ac{E2V} generation and introduce \textit{E2V-Bench}, a novel benchmark for generating pedagogically meaningful images from arithmetic equations, together with automatic evaluation metrics for assessing the quality of generated visuals.
To ground our benchmark design in instructional practice, we manually analyzed educational visuals from six educational resources and identified four recurring visual types. We further interviewed ten primary-school mathematics teachers to refine and validate these design choices.

Using \textit{E2V-Bench}, we evaluate nine representative \ac{T2I} models across five model families. 
While several models show non-trivial performance, as shown in Fig.~\ref{fig:fig0}, even the strongest models exhibit systematic failures in generating pedagogically faithful visuals.
In contrast, a structured DSL-based reference method achieves near-perfect accuracy, suggesting that while the task is well specified within our benchmark scope, current \ac{T2I} models still struggle to produce pedagogically faithful visual representations.
Motivated by these findings, we explore benchmark-guided enhancement strategies, including prompt refinement, regeneration, and rejection-sampling supervised fine-tuning with curated training data.
Our results show that these strategies yield notable improvements for representative models, highlighting directions for making visual generation more reliable in educational settings. Our contributions are:

\noindent \circled{1} \textit{E2V-Bench}, the first benchmark dedicated to equation-to-visual generation in early arithmetic education, grounded in educational resources and teacher feedback.

\noindent \circled{2} A systematic evaluation of representative \ac{T2I} models showing that models struggle to generate pedagogically faithful visuals for equations.

\noindent \circled{3} A high-quality E2V training dataset and benchmark-guided enhancement strategies that improve representative models on E2V generation.

\section{Related Work}

\paragraph{Visuals in Teaching Primary-School Arithmetic}
Visuals are widely recognized as beneficial for learning arithmetic skills~\citep{kaitera2022developing,boonen2016s}.
Well-designed visuals make mathematical ideas more accessible~\citep{MAYER200285, evagorou2015role, small2025eyes}, foster student engagement~\citep{cooper2018benefits}, and support more efficient learning~\citep{arcavi2003role}. Various pedagogical designs have been proposed to teach arithmetic. 
For instance, the bar model~\citep{hoven2007singapore} expresses numerical relations through proportional bars, and has been shown to enhance children's problem-solving ability~\citep{Osman2018} and use of effective cognitive strategies~\citep{morin2017use}. 
The Noyon framework depicts mathematical problems with modular iconic components, offering a flexible yet structured approach to visualize mathematical reasoning~\citep{saquib2021constructing}. 
These approaches underscore the importance of carefully designed visuals in arithmetic education.

\paragraph{Pedagogical Visual Retrieval and Generation}
Despite their benefits, producing educational visuals requires considerable effort from teachers~\citep{xu2021procedural}, motivating automation of visual retrieval and generation.
Prior work has explored image retrieval for textbook development~\citep{singh-etal-2023-enhancing}, generating multiple-choice questions from semantic representations~\citep{singh2019automatic}, and combining multimodal information for question generation~\citep{luo-etal-2024-chain}.
However, these methods do not focus on visualizing mathematical concepts for learning. 
Early efforts such as VisualMath~\citep{dwivedi2017visualmath} rendered simple word problems with pre-existing images, but addressed only basic operations (+, -) and lacked pedagogical validation. 
More recently, Math2Visual~\citep{wang-etal-2025-generating-pedagogically} generates visuals from math word problems with stronger educational grounding, but its reliance on fixed templates limits flexibility in representing arithmetic concepts. 
These works highlight progress toward automation but underscore the absence of evaluations and methods for tuning models to generate educational visuals.

\section{Equation-to-Visual Generation Task} \label{sec:task_define}
We define the task of \ac{E2V} generation as producing visuals that faithfully represent the arithmetic concepts encoded in an equation. 
Rather than reproducing equations as text, generated visuals should depict quantities through concrete objects and spatial arrangements that support learners' intuitive understanding of arithmetic relationships: a form of visual scaffolding commonly used in early mathematics instruction~\citep{kaitera2022developing,boonen2016s}.

To ground this task in authentic educational practice, we began by examining how arithmetic equations are visualized in existing instructional materials. Two researchers manually collected 189 visuals from six widely used educational resources, including three textbooks and three online platforms (full list in App.~\ref{sec:thematic_analysis}). 
After excluding seven decorative visuals that did not correspond to arithmetic equations, we conducted a two-stage thematic analysis on the remaining 182 visuals: an exploratory review of 50 visuals to identify recurring patterns, followed by coding the remaining 132 visuals using the identified categories (details are in App.~\ref{sec:thematic_analysis}). This analysis revealed four recurring visual types for representing arithmetic equations:

\noindent \textbf{(1) Color-based:}
Operands are distinguished by assigning distinct colors to otherwise identical objects (e.g., red vs.\ green apples). Color serves as the primary grouping cue, a perceptual feature that children readily use for grouping~\citep{harris1970effects}.

\noindent \textbf{(2) Object-based:}
Operands are distinguished by using different object types, with object category serving as the primary grouping cue. Prior work shows that children can readily organize objects by type~\citep{bornstein2010development}.

\noindent \textbf{(3) Spatial-based:}
Operands are distinguished by placing object groups in different spatial positions (e.g., left vs.\ right). Spatial separation serves as the primary grouping cue, which children naturally use to organize visual information~\citep{quinn2008young}.

\noindent \textbf{(4) Container-based:}
Operands are distinguished by containment relations, such as placing one group inside a container and another outside, or placing groups into different containers. Children can reason about containment~\citep{casasola2003six}, making containers natural for grouping.

Based on this analysis, we define the \ac{E2V} task as generating accurate visuals for an equation under a specified visual type. 
The equation determines the quantities and arithmetic relation, while the visual type determines how operands are visually grouped. 
We formalize each equation--visual type pair with a \ac{DSL},
\[
t\{g_i\{\texttt{item}, \texttt{count}, a_t\}\}_{i=1}^{n},
\]
where $t$ denotes the visual type, $g_i$ an operand group, \texttt{item}/\texttt{count} the object category and quantity within the group, and $a_t$ the grouping attribute, such as \texttt{color}, \texttt{position}, or \texttt{container}. For object-based visuals, the object category itself serves as the grouping cue, so no additional grouping attribute is required.
This \ac{DSL} serves as the reference semantics for generation and evaluation; details and examples are in App.~\ref{sec:dsl_detail}.
We also observed that educational visuals commonly appear in both cartoon and realistic styles, which may influence how learners perceive generated content.
Although our task formulation supports both styles, we focus the main analysis on realistic-style visuals and report cartoon-style results in App.~\ref{sec:cartoon_result}.

\subsection{Evaluative Study with Teachers} \label{sec:evaluative_study}
To assess the pedagogical value of the proposed visual types, we conducted an evaluative study with ten experienced primary school math teachers (Grades 1-3; demographics in Tab.~\ref{tab:tab5}). Teachers reviewed visuals following our four visual types, covering both cartoon and realistic style designs across the four basic arithmetic operations, and rated their pedagogical usefulness on a 7-point Likert scale. Study details are provided in App.~\ref{sec:teacher_study}.

\noindent \textbf{The four visual types align with current classroom practice.} Teachers consistently reported that the four visual types reflect visuals commonly used in Grade 1-3 classrooms (Tab.~\ref{tab:tab10}). All participants indicated prior experience with similar visuals, suggesting that our designs align well with existing instructional practices. Across evaluation criteria, teachers gave uniformly high ratings to visuals generated using the proposed visual types (Tab.~\ref{tab:tab6}), indicating teachers find visuals following our proposed visual types pedagogically valuable and effective for supporting arithmetic learning.

\noindent \textbf{Visual type preferences.}
Teachers' preferences for visual types varied by operation (Tab.~\ref{tab:tab9}). Container-based visuals were favored for multiplication and division because explicit grouping helps convey the underlying operations. Color-based visuals were commonly linked to addition, where color cues distinguish addends while showing their combination. Spatial-based visuals were often considered useful for subtraction, as spatial separation supports the idea of removing quantities, and also for grouping in multiplication and division. Object-based visuals were viewed as more cognitively demanding and better suited for students who already have a basic understanding of the operations.

\section{E2V-Bench: Dataset and Metrics}

Although the broader E2V goal is defined over equations, directly prompting existing T2I models with symbolic expressions is ineffective, as models tend to generate stylized renderings of equations rather than object-based visuals that convey quantity and structure (App.~\ref{sec:naive_gen}). This limitation arises because current T2I models are designed for natural language inputs rather than symbolic expressions. To address this gap, E2V-Bench standardizes the intermediate equation-to-visual-description step: each equation is verbalized into a \ac{VD} that specifies the intended quantities, objects, and grouping cue, and models are evaluated on whether they can faithfully render this VD. 
This formulation evaluates image generation under a standardized equation-to-\ac{VD} layer, isolating it from open-ended verbalization while still reflecting a practical equation-to-visual pipeline.

\begin{table}
\centering
\scriptsize
\setlength{\tabcolsep}{2.5pt}
\begin{tabular}{p{1.5cm}p{1.8cm}|>{\centering\arraybackslash}m{1.9cm}>{\centering\arraybackslash}m{1.7cm}}
\toprule
\textbf{Category} & \textbf{Model} & 
\textbf{Quantity Acc.} & 
\textbf{Overall Acc.} \\
\midrule
DSL-based
& DSL-based (ours)
& $\mathbf{98.3 \pm 1.4}$ & $\mathbf{98.3 \pm 1.4}$ \\
\midrule
\makecell[l]{Prompt\\Refinement}
& PAE        
& $1.0 \pm 1.1$ & $1.0 \pm 1.1$ \\
\midrule
\multirow{2}{=}{Closed-source}
& Recraft-v3   
& $6.0 \pm 2.7$ & $4.7 \pm 2.4$ \\
& GPT-Image-1 
& $27.0 \pm 5.0$ & $21.3 \pm 4.6$ \\
\midrule
\multirow{3}{=}{Diffusion-based}
& SD-3.5-large
& $6.7 \pm 2.9$ & $3.3 \pm 2.1$ \\
& Flux.1-dev 
& $8.7 \pm 3.2$ & $6.3 \pm 2.8$ \\
& \cellcolor{hj_gray!30} Flux.1-dev (SFT)
& \cellcolor{hj_gray!30} $8.0 \pm 3.1$
& \cellcolor{hj_gray!30} $5.3 \pm 2.6$ \\
\midrule
\multirow{4}{=}{Layout-to-image}
& Blueprint   
& $2.7 \pm 1.8$ & $1.3 \pm 1.3$ \\
& LMD         
& $22.4 \pm 4.8$ & $12.0 \pm 3.7$ \\
& \cellcolor{hj_gray!30} LMD (PR)
& \cellcolor{hj_gray!30} $26.7 \pm 5.1$
& \cellcolor{hj_gray!30} $14.7 \pm 4.1$ \\
& \cellcolor{hj_gray!70} LMD (PR+Regen)
& \cellcolor{hj_gray!70} $26.7 \pm 5.0$
& \cellcolor{hj_gray!70} $15.3 \pm 4.1$ \\
\midrule
\multirow{3}{=}{Transformer-based}
& Show-o2 
& $4.3 \pm 2.3$ & $2.3 \pm 1.7$ \\
& Bagel     
& $8.0 \pm 3.1$ & $5.7 \pm 2.6$ \\
& \cellcolor{hj_gray!30} Bagel (SFT)
& \cellcolor{hj_gray!30} $8.0 \pm 3.1$
& \cellcolor{hj_gray!30} $6.3 \pm 2.8$ \\
& \cellcolor{hj_gray!70} Bagel (RSFT)
& \cellcolor{hj_gray!70} $16.7 \pm 4.3 $ 
& \cellcolor{hj_gray!70} $14.7 \pm 4.0$\\

\bottomrule
\end{tabular}
\caption{\textbf{Evaluation of T2I models and enhanced variants on E2V-Bench.} We report quantity and overall accuracy with 95\% confidence interval half-widths. PR denotes prompt refinement, Regen denotes regeneration. SD-3.5-large denotes Stable Diffusion-3.5-large.}
\label{tab:tab1_realistic}
\vspace{-5pt}
\end{table}

\subsection{Benchmark Construction}
\begin{figure}[t]
    \centering
    \includegraphics[width=\linewidth]{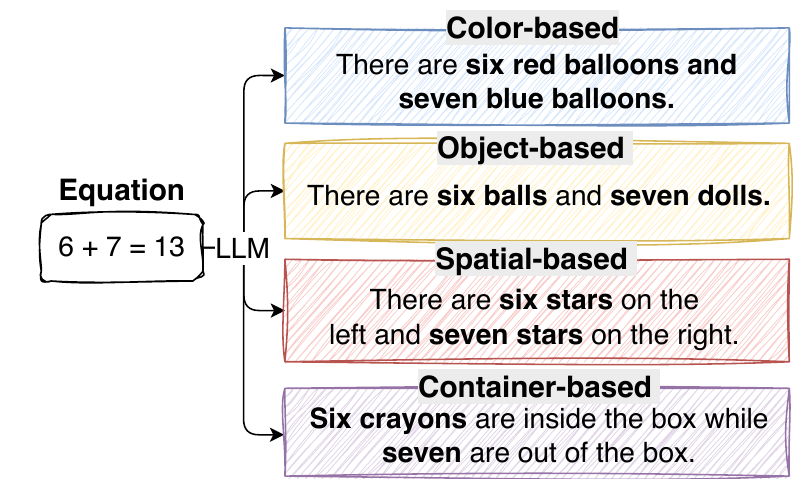}
    \caption{\textbf{Standardized equation-to-VD layer used in E2V-Bench.} For each equation, we use an LLM to generate four VDs, one for each visual type. These VDs are used as prompts for \ac{T2I} models to produce visuals. }
    \label{fig:fig1}
    \vspace{-15pt}
\end{figure}

We construct \ac{E2V-Bench} in two phases: equation generation and \ac{VD} generation (Fig.~\ref{fig:fig1}). 
We enumerate arithmetic equations over the four basic operations (+, --, $\times$, $\div$), restricting quantities to at most 20 for primary-school suitability, yielding 371 unique equations. 
Using in-context learning with Gemini-2.5-Flash~\citep{googleGeminiFlash}, we generate four \ac{VD}s per equation, one for each visual type, guided by eight manually crafted examples per type; we further use Gemini-2.5-Flash to annotate each \ac{VD} with its corresponding \ac{DSL} for evaluation.

Two researchers verified 120 sampled \ac{VD}--\ac{DSL} pairs (30 per visual type), confirming alignment with the intended equations and visual types.
Pedagogical value was further validated by ten primary-school teachers, who evaluated 32 generated \ac{VD}s and rated them as suitable for classroom use (App.~\ref{sec:teacher_result}). The dataset contains 1,484 \ac{VD}s: 1,184 for training and 300 for testing (statistics in Tab.~\ref{tab:tab15}), evenly balanced across the four visual types.

\subsection{Evaluation Metrics}\label{sec:metric}
We evaluate model outputs using two criteria informed by prior work and interviews with ten primary-school mathematics teachers.
Both criteria compare each generated visual against the \ac{DSL} induced by its equation and visual type.

\noindent\textbf{Quantity Accuracy} measures whether the number of visual instances in each operand group matches the count specified by the \ac{DSL}. 
A visual is considered quantity accurate only if all operand groups have correct counts. 
To compute this automatically, we adapt CountGD~\citep{AminiNaieni24}, which detects instances of each target object type in the visual and returns their total count.

\noindent\textbf{Overall Accuracy} measures whether a visual correctly satisfies all semantic and structural constraints induced by the \ac{DSL}. 
While recent benchmarks often rely on vision-language models as judges, we intentionally avoid this approach due to their unreliability in object counting and visual verification~\citep{choudhury2025can}. 
To ensure precise and reproducible evaluation, we adopt a deterministic evaluation pipeline implemented with OpenCV and Scikit-Learn.
For each generated image, we first detect object instances using CountGD, then apply verification rules conditioned on the visual type. 
For example, in color-based visuals, our method checks whether the detected object colors align with the expected color distribution defined by the corresponding semantic representation. 
Implementation details are in App.~\ref{sec:auto_metric}.

Both criteria are critical in educational settings, where learning depends on the correctness of instructional information~\citep{metzger2003college,goldin2001systems}.
To validate alignment with human judgment, we randomly sampled 216 visuals generated by the nine models from the \ac{E2V-Bench} test set and had two researchers independently annotate each visual.
They disagreed on only two cases, resolved through discussion.
Cohen's $\kappa$ between human annotations and automatic metrics was 0.96 for both criteria, indicating strong agreement; agreement rates by quantity range and visual type are reported in App.~\ref{sec:metric_validation_de}.
To further assess pedagogical validity, we collected expert feedback from ten primary-school teachers during the user study; all teachers agreed that the two criteria adequately capture visual quality for teaching arithmetic skills (details in App.~\ref{sec:teacher_study}).

\section{Benchmarking T2I Models}

We evaluate nine \ac{T2I} models on E2V-Bench, covering five representative model families: closed-source models (GPT-Image-1, Recraft-v3), which are proprietary general-purpose image generation systems; diffusion-based models (Flux.1-dev, Stable Diffusion-3.5-large), which synthesize images through iterative denoising; layout-to-image models (LMD, Blueprint), which first generate text-based bounding box layouts and then condition a diffusion model on them; transformer-based models (Show-o2, Bagel), which integrate multimodal understanding and generation within a single transformer architecture; and a prompt-refinement model (PAE), which iteratively rewrites prompts for \ac{T2I} models. 
Model references are provided in Tab.~\ref{tab:tab1_cartoon}.
We additionally include a \ac{DSL}-based method inspired by Math2Visual~\citep{wang-etal-2025-generating-pedagogically}, which renders visuals from the \ac{DSL} using icon assets. Given a \ac{DSL} specification, the renderer selects icons according to the object category, places them according to the visual type and grouping attributes, and recolors icons for color-based visuals.
Because it relies on manually specified rendering rules and directly receives the target specification, we treat it as a reference rather than as a comparable \ac{T2I} baseline; details are in App.~\ref{sec:dsl_gen_detail}.

\vspace{-1pt}
\begin{figure}[t]
    \centering

    \includegraphics[width=\linewidth]{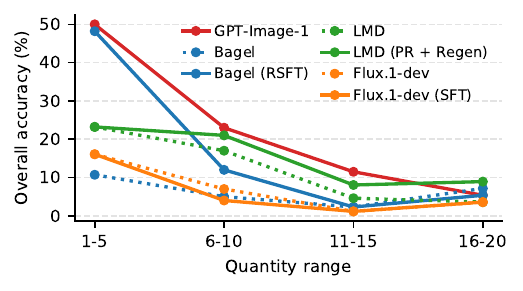}

    \caption{\textbf{Overall accuracy across quantity ranges.} Model performance generally declines as object counts increase, highlighting quantity control as a major challenge in E2V generation.}

    \label{fig:fig5}
    \vspace{-15pt}
\end{figure}
\paragraph{Performance Across Model Families.}

Tab.~\ref{tab:tab1_realistic} reports the evaluation results with 95\% confidence interval half-widths estimated from 10,000 bootstrap resamples.
Among the nine \ac{T2I} models, GPT-Image-1 achieves the best performance, while the other closed-source model, Recraft-v3, performs substantially worse.
Open-source models exhibit substantial variation across model families. 
Diffusion-based models perform poorly overall: Stable Diffusion-3.5-large achieves low overall accuracy, while Flux.1-dev shows modest improvements but still struggles to reliably control object counts.
Layout-to-image models show mixed performance. 
Blueprint performs worst across most metrics, whereas LMD substantially outperforms other open-source models, suggesting that explicit layout conditioning benefits numerical grounding. 
Transformer-based models exhibit variable behavior: Bagel surpasses Show-o2 in both quantity and overall accuracy.
Finally, PAE records near-zero overall accuracy, confirming that prompt refinement alone is ineffective at generating images with correct object counts, consistent with prior findings~\citep{cao2025text}.
The \ac{DSL}-based method reaches 98.3\% overall accuracy, suggesting that \ac{E2V} is largely solvable with explicit structure, while current \ac{T2I} models still struggle on this task.
We therefore further analyze the sources of this gap in \ac{T2I} models, focusing on how performance varies across quantity ranges and visual types.

\vspace{-1pt}

\paragraph{Variation Across Quantities and Visual Types.} \label{sec:variation_quantity}

Across models, overall accuracy declines sharply as object counts increase (Figure 3). The steepest drop occurs when moving from 1–5 objects to 6–10 objects, with several stronger models falling from around 50\% to roughly 20–25\%.
To better understand this drop, we conducted an error analysis on 208 randomly sampled visuals from four representative models. Error cases were manually coded into quantity error (90.4\%), structural error (16.8\%), object mismatch (11.1\%), and other failures (2.9\%), with each sample allowed to receive multiple error labels. These results suggest that the performance drop is driven primarily by quantity errors; details are provided in App.~\ref{app:error_analysis}.
Performance also varies by visual type (Fig.~\ref{fig:fig6}): GPT-Image-1 shows the most balanced profile, particularly outperforming other models in the color visual type.  LMD performs comparably to GPT-Image-1 in the object, spatial, and container visual types. 
Bagel emerges as the second-best performer on container and color-based visuals, indicating its potential for further improvement. 
The LMD (PR+Regen), Bagel (RSFT) and Flux.1-dev (SFT) are enhanced versions of LMD, Bagel and Flux.1-dev; we will discuss these in Sec.~\ref{sec:improve_model}.

\vspace{-1pt}

\paragraph{Key Insights and Open-source Potential.}

Overall, GPT-Image-1 delivers the highest-quality outputs but cannot serve as a reproducible research baseline due to its proprietary nature. Among open-source models, LMD, Flux.1-dev, and Bagel stand out as the strongest representatives and demonstrate potential for further improvement.

\section{Enhancing E2V Generation}\label{sec:improve_model}

Motivated by our error analysis (Sec.~\ref{sec:variation_quantity}), which identified quantity errors as the dominant failure mode and structural issues as a secondary source of errors, we use E2V-Bench to explore diagnostic, benchmark-guided strategies for improving representative T2I models: LMD, Bagel, and Flux.1-dev.

\subsection{Layout-to-Image Model}

\begin{figure}[t]
    \centering

    \includegraphics[width=\linewidth]{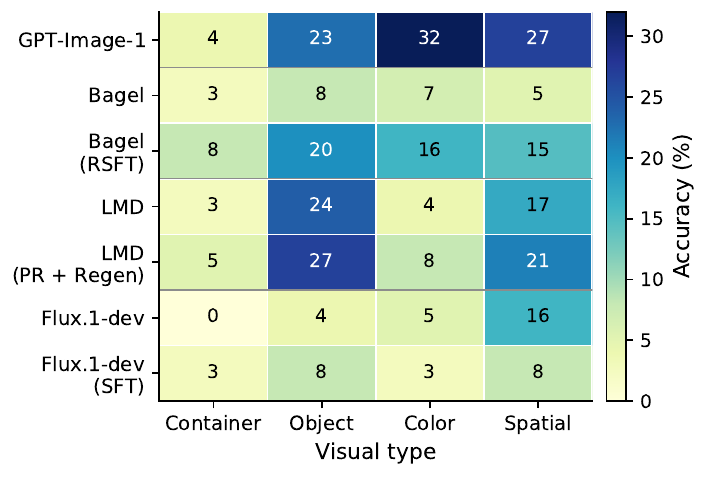}

    \caption{\textbf{Overall accuracy across visual types.} Performance varies by visual type, showing that grouping cues introduce different levels of difficulty for T2I models.}

    \label{fig:fig6}
    \vspace{-15pt}
\end{figure}

LMD\footnote{All LMD experiments use GPT-4o for layout generation, the state-of-the-art multimodal model at the time~\citep{cao2025preliminary, ramachandran2025well}.} is a representative layout-to-image framework.
To understand why it underperforms on E2V-Bench, we analyzed the layouts produced by LMD.
Two researchers conducted a thematic analysis of 50 generated layouts. The analysis proceeded in two phases: an initial exploration of 20 layouts to identify recurring error types, followed by a systematic evaluation of the remaining 30 layouts using these categories. The two researchers initially disagreed on four cases, which were resolved through discussion, resulting in three common error types:
(1) Quantity error: the number of generated objects does not match the input \ac{VD}; 
(2) Boundary error: bounding boxes extend beyond the image boundary; and 
(3) Format error: invalid JSON output that cannot be parsed.
Grounded in these findings, we developed rule-based scripts to automatically detect each error type. Tab.~\ref{tab:tab3} reports the frequencies of these errors across 300 samples from the \ac{E2V-Bench} test set.
These results suggest that many of LMD's failures arise during layout generation, before final image synthesis.
Based on this analysis, we explore the following two strategies to improve layout generation:

\paragraph{Prompt refinement.}

Motivated by our analysis showing that many failures originate during layout generation, we refine the prompt to better guide scene-level planning before box prediction. Rather than asking the model to directly output bounding boxes, the revised prompt frames layout generation as a \emph{scene design} task: the model is first asked to plan a coherent scene that can accommodate all required objects, and then to generate bounding boxes consistent with that scene. Concretely, the prompt instructs the model to use a plausible setting, keep all target objects clearly visible and sufficiently large, place boxes within image boundaries, avoid unnecessary overlap, and organize objects in spatially meaningful ways that support counting and grouping. It also encourages simple configurations, such as grouping related objects and adopting top-down views in crowded scenes, to reduce ambiguity in the final layout. In addition, we provide 15 manually curated in-context examples for \ac{E2V} that demonstrate layouts satisfying these spatial and pedagogical constraints. Original and refined prompts are provided in App.~\ref{sec:prompt_refine}.

\paragraph{Regeneration.}

We further introduce an automatic regeneration loop that uses rule-based error detectors derived from layout error analysis. These detectors check whether the generated layout contains the correct number of objects (quantity error), whether any bounding boxes extend beyond the image boundary (boundary error), and whether the output follows the required JSON format (format error). After an initial layout is generated, we apply these detectors to identify potential errors. If any error is found, we construct a targeted feedback prompt that explicitly describes the detected issue (e.g., ``the previously generated layout contains an incorrect number of objects'') and provide the previous layout as context to guide correction. To limit computation and avoid over-correction, we allow up to two regeneration attempts per layout.

\paragraph{Results.}

Prompt refinement and regeneration lead to reductions across all layout error types, as detected by our scripts, and yield corresponding improvements in downstream visual accuracy.
Tab.~\ref{tab:tab3} shows that prompt refinement notably reduces layout errors, with further gains from regeneration.
These improvements translate into better end-task performance on the \ac{E2V-Bench} test set (Tab.~\ref{tab:tab1_realistic}), where overall accuracy increases from 12.0\% to 15.3\% with prompt refinement and regeneration.

\begin{table}[t]
\centering
\scalebox{0.66}{
\setlength{\tabcolsep}{3pt}
\begin{tabular}{lccc}
\toprule
\textbf{Method} & \textbf{Quant Err. (\%)} & \textbf{Bound Err. (\%)} & \textbf{Format Err. (\%)} \\
\midrule
LMD             & 41.7 & 18.7 & 2.0 \\
\rowcolor{hj_gray!30}LMD (PR)          & 27.0 & 22.3 & \textbf{0.0} \\
\rowcolor{hj_gray!70}LMD (PR+Regen)    & \textbf{21.0} & \textbf{6.7} & \textbf{0.0} \\
\bottomrule
\end{tabular}
}
\caption{\textbf{Layout-generation error rates for LMD variants.}
Prompt refinement and regeneration reduce quantity, boundary, and format errors, suggesting that many downstream failures originate from layout planning.}
\label{tab:tab3}
\vspace{-10pt}
\end{table}

\subsection{Diffusion and Transformer-based Models}

Next, we explore whether supervised fine-tuning on a curated dataset can improve T2I models, including a diffusion model (Flux.1-dev) and a transformer-based model (Bagel).

\paragraph{Training Dataset Construction.}

We curate training data with a generation-and-filtering pipeline:
generating 5,920 \ac{VD}s from the \ac{E2V-Bench} training equations and using GPT-Image-1, the best-performing model in our benchmark, to generate corresponding visuals.
We then filter the generated images using our automatic metrics (see Sec.~\ref{sec:metric}) and retain those with no errors (overall accuracy of one), resulting in 1,055 image--\ac{VD} pairs (Tab.~\ref{tab:tab17}).
We fine-tuned both Flux.1-dev and Bagel for 10k steps (details are in App.~\ref{sec:flux_sft} and ~\ref{sec:bagel_sft}).

\begin{figure}
    \centering

    \includegraphics[width=\linewidth]{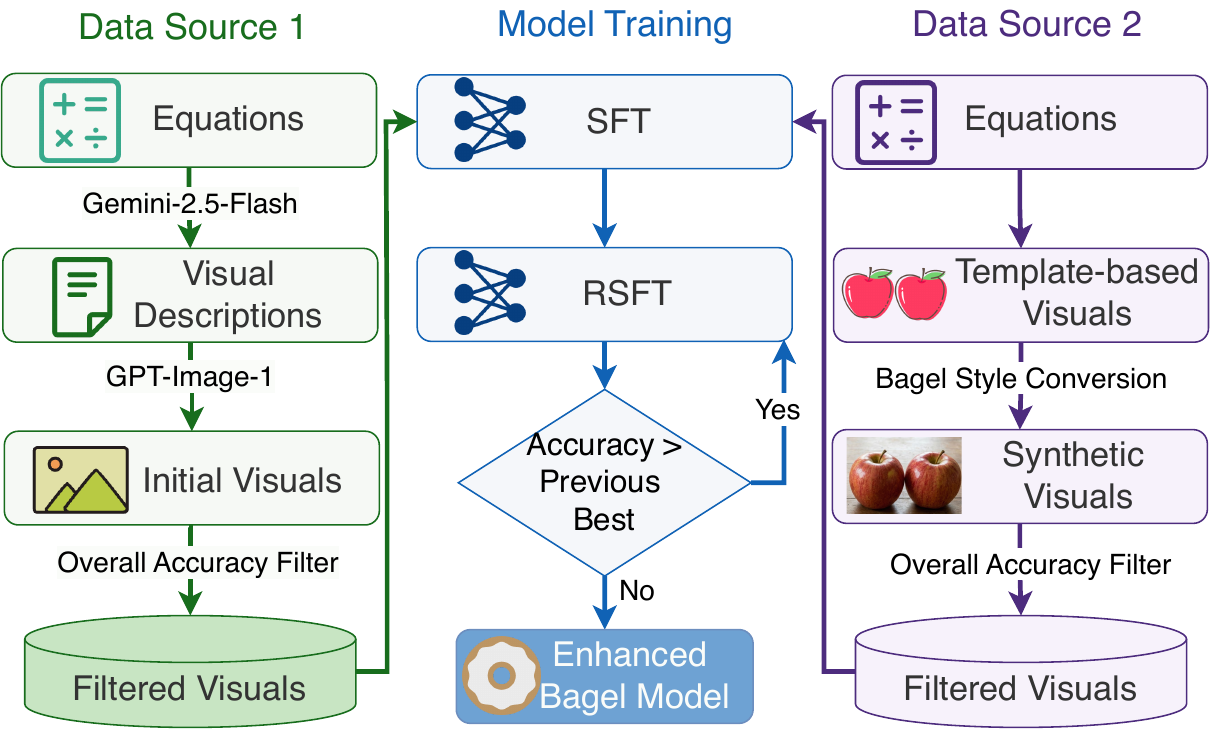}

    \caption{\textbf{Bagel training pipeline.}
The training process begins with SFT on filtered GPT-Image-1 generated data, continues with SFT on a mixed dataset of synthetic data and GPT-Image-1 data, and concludes with iterative rejection-sampling supervised fine-tuning.}
    \label{fig:fig10}
    \vspace{-15pt}
\end{figure}
\paragraph{Results.}
Tab.~\ref{tab:tab1_realistic} reveals divergent trends: fine-tuning slightly improves the overall accuracy of Bagel from 5.7\% to 6.3\%, but degrades the overall accuracy of Flux.1-dev.
We hypothesize that Bagel, an LLM-initialized unified multimodal model trained on interleaved data, retains stronger numerical reasoning capabilities than Flux's purely generative diffusion architecture, making it more robust for further enhancement.
This is consistent with prior findings on the limitations of diffusion models in numerical reasoning~\citep{kajic2024evaluating}.

\subsubsection{Further Enhancements to Bagel}
 
\begin{figure}[t]
    \centering
    \includegraphics[width=\linewidth,
    ]{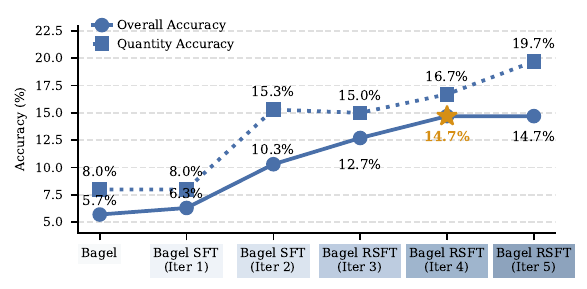}
    \caption{\textbf{Bagel performance across training stages.} Overall accuracy improves after synthetic augmentation and RSFT, showing gains from structured data curation. Star indicates the best-performing checkpoint.}
    \label{fig:fig9}
    \vspace{-15pt}
\end{figure}

We further improve Bagel's performance with synthetic augmentation and iterative refinement (Fig.~\ref{fig:fig10}).
We constructed a synthetic dataset using template-based scripts and icon assets, with equations drawn from the \ac{E2V-Bench} training split. 
As the synthetic images differ from Bagel-generated outputs in their pixel-level distribution, we adapted them using Bagel's style transfer module and applied metric-based filtering to retain higher-quality 3,017 examples (details in App.~\ref{appendix:syn_data}).
Following prior work~\citep{adarsh-etal-2025-siked}, we mixed the filtered synthetic data with the original training set and performed SFT from the previously fine-tuned checkpoint. 
We adopt this strategy because mixing GPT and synthetic data in the initial fine-tuning stage led to lower performance (App.~\ref{sec:bagel_ablation}), likely because the GPT-Image-1 data provides a stronger initial generation prior.
As shown in Fig.~\ref{fig:fig9}, the second SFT stage further improves overall accuracy.

We further applied rejection-sampling supervised fine-tuning (RSFT), generating ten candidate outputs per training \ac{VD} and retaining samples judged correct by overall accuracy, with at most three visuals preserved per \ac{VD} to avoid over-representing easy cases (see Fig.~\ref{fig:fig10}). Each RSFT round was trained for 10k steps, and training was stopped once performance no longer improved. As shown in Fig.~\ref{fig:fig9}, RSFT improves overall accuracy to 14.7\%.
Further RSFT iterations do not yield additional improvement, potentially due to reduced sample diversity and increased focus on easy cases.

To verify that the gains were not simply due to additional training and did not merely reflect closer alignment to the automatic evaluation metric, we conducted two additional analyses, with component-level ablations reported in App.~\ref{sec:bagel_ablation}. Fine-tuning Bagel on the original training data alone for 20k, 30k, and 40k steps yielded overall accuracies of 5.7\%, 6.3\%, and 7.7\%, respectively, indicating that increased training alone does not explain the improvement. 
We further randomly sampled 300 visuals from GPT-Image-1, Bagel (RSFT), and LMD (PR+Regen), and two researchers independently annotated them for Overall Accuracy and Quantity Accuracy. They disagreed on three cases, which were resolved through discussion. The automatic judgments showed high agreement with human labels, with exact agreement rates of 0.98 and 0.94 and Cohen's kappa values of 0.92 and 0.82 for Overall Accuracy and Quantity Accuracy, respectively, suggesting that the gains reflect improved arithmetic reasoning and visual grounding rather than overfitting to the evaluation metric.

\vspace{-1.8pt}
\subsection{Analysis and Real-World Validation}
Fig.~\ref{fig:fig5} and Fig.~\ref{fig:fig6} summarize model performance after enhancements. 
In Fig.~\ref{fig:fig5}, Bagel (RSFT) achieves accuracy similar to GPT-Image-1 in the low-quantity range, but its performance declines rapidly as object counts increase.
In contrast, LMD (PR+Regen) attains lower accuracy in low-quantity settings but exhibits more stable performance as quantities grow, suggesting that layout priors help maintain robustness under higher object counts.

Across all visual types (Fig.~\ref{fig:fig6}), Bagel (RSFT) improves over the original Bagel, performing best on container-based and second best on color-based visuals. 
LMD (PR+Regen) improves across all visual types, performing best on object-based and second-best on spatial-based visuals, indicating that layout-based priors are effective for generating correct objects and precise spatial arrangements.

\begin{figure}[t]
    \centering
    \includegraphics[width=\linewidth,
    ]{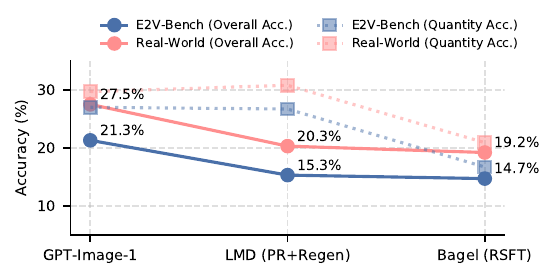}

    \caption{\textbf{Human evaluation on the real-world educational visuals and \ac{E2V-Bench}.} Performance trends on real-world dataset mirror those on \ac{E2V-Bench}, suggesting that the \ac{E2V-Bench} captures core pedagogical challenges present in authentic educational visuals.}
    \label{fig:fig11}
    \vspace{-15pt}
\end{figure}

To validate model performance in authentic educational settings, we conducted a human evaluation on a manually curated dataset of real-world educational visuals paired with \ac{VD}s, collected from textbooks and online educational platforms (§\ref{sec:task_define}).
The resulting dataset covers the four basic operations and comprises 182 visuals.
Using these \ac{VD}s as prompts, we evaluated three representative models: GPT-Image-1, LMD (PR+Regen), and Bagel (RSFT), by generating 546 visuals.
Two researchers independently compared each generated visual against the ground-truth visual and assigned binary judgments for Quantity and Overall Accuracy using the criteria defined in Sec.~\ref{sec:metric}, resolving three initial disagreements through discussion (details in App.~\ref{sec:human_eval}).
As shown in Fig.~\ref{fig:fig11}, performance trends on this real-world dataset closely mirror those observed on \ac{E2V-Bench}, suggesting that the benchmark captures core pedagogical challenges present in authentic educational visuals.

\vspace{-1.8pt}
\section{Conclusion}
\vspace{-1.9pt}

We introduced the equation-to-visual generation task and \ac{E2V-Bench}, a benchmark to support the study of generating pedagogically meaningful visuals for arithmetic education. Grounded in primary-school math teachers' insights and educational resources, E2V-Bench defines four visual types and includes automatic evaluation metrics.
Our experiments show that current text-to-image models struggle with this task, while targeted strategies guided by E2V-Bench and supported by curated training data can partially reduce this gap.
Together, these contributions establish a foundation for equation-to-visual generation in education and outline pathways for improving model performance.

\section*{Limitations}

\paragraph{Scope of Representation}
Our method is scoped to early arithmetic education (Grades 1–3), focusing on the four basic operations and quantities below 20. This design choice reflects common classroom practice and enables a precise definition of pedagogical correctness. By constraining the scope, we are able to construct a well-grounded benchmark and reliable evaluation framework. Nonetheless, this focus limits direct applicability to larger quantities, more complex operations, and more advanced mathematical concepts, such as fractions, ratios, or multi-step equations. Extending equation-to-visual generation beyond early arithmetic remains an important direction for future work. At the same time, given that current models already struggle in this elementary setting, reliable generation for foundational arithmetic is a necessary first step.

\paragraph{Coverage of Visual Types}

Our method models four pedagogically grounded visual types distilled from textbook analysis and validated through teacher feedback. These visual types capture recurring and widely used educational visual patterns in early arithmetic, enabling the task to be formalized and evaluated in a principled manner. 
However, they do not exhaust the full space of educational visual representations. 
Future work could expand the visual taxonomy to cover a broader range of pedagogical designs.

\paragraph{Focus on Foundational Visual Accuracy}

Our evaluation focuses on visual correctness, specifically quantity and structural alignment with the intended arithmetic concepts, rather than downstream learning outcomes. This emphasis is deliberate: accurate visual representation is a prerequisite for any instructional use, and our experiments show that current \ac{T2I} models still exhibit low overall accuracy on this task (with the strongest models achieving around 20\% overall accuracy). Because incorrect quantities or structures can mislead learners, generated visuals should not be used in educational settings without human verification. Given this gap, establishing reliable generation and evaluation of correct visuals is a necessary first step before studying higher-level educational effects. Once model accuracy reaches a reliably high level, future work can incorporate human-centered studies to examine learning outcomes, pedagogical impact, and practical teacher-facing deployment.
\section*{Acknowledgements}
This project was supported by ETH AI Center Doctoral Fellowships to Junling Wang and Boqi Chen, ETH AI Center Postdoctoral Fellowships to Heejin Do and Mubashara Akhtar, the Swiss AI Large Grant SCR1089274, and Swiss AI Small Grant \#63, and a project grant from the SNSF (Grant No. 10009282).
The authors also thank the reviewers, members of the LRE Lab and PEACH Lab at ETH Zurich, and the participants in the human evaluation experiments for their valuable feedback and contributions.

\bibliography{custom}

\appendix
\label{sec:appendix}

\section{Details of Thematic Analysis of Visual Types} \label{sec:thematic_analysis}
\subsection{Procedure}
We conducted a thematic analysis to identify recurring visual types from visuals collected across six educational sources, including three textbooks~\citep{southsudan_math2_2018,cotton2021oxfordmaths1,mosely2021cambridge2} and three online platforms~\citep{accessim,mathematicsmonsterMathematicsMonster,fun2dolabs}.
In total, we collected 189 visuals; seven were decorative and did not correspond to their associated equations, and were therefore excluded, leaving 182 visuals.
In the initial phase, we sampled 50 visuals to identify recurring themes. Through iterative discussion, two researchers identified and consolidated four distinct visual types. Coding was performed collaboratively, with a focus on refining the categories through close reading and comparison.

In the systematic evaluation phase, the same two researchers manually analyzed the remaining 132 visuals. This broader analysis allowed us to assess the distribution of problems across the four identified visual types.

\subsection{Results}
The results of the systematic evaluation are presented in Tab.~\ref{tab:tab14}. The distribution of examples is relatively balanced across spatial-based (54), object-based (50), container-based (49), and color-based (29) visual types, suggesting that the identified visual types can effectively categorize visuals from educational sources.

\begin{table}[h]

\centering
\small
\begin{tabular}{lcccc}
\toprule
\textbf{Category} & \textbf{Spatial} & \textbf{Object} & \textbf{Container} & \textbf{Color}  \\
\midrule
\textbf{Count} & 54 & 50 & 49 & 29 \\
\bottomrule
\end{tabular}
\caption{Distribution of examples across visual types.}
\label{tab:tab14}
\end{table}


\section{Dataset Details}
\subsection{Statistics of E2V-Bench Dataset and Training Dataset}
The statistics of the \ac{E2V-Bench} dataset are shown in Tab.~\ref{tab:tab15}, while those of our curated ground-truth training dataset are presented in Tab.~\ref{tab:tab17}.

\begin{table}[h]

\centering
\scriptsize
\begin{tabular}{llrrr}
\toprule
\textbf{Category} & \textbf{Sub-category} & \textbf{Train} & \textbf{Test} & \textbf{Total} \\
\midrule
\multirow{4}{*}{Visual Type}

  & Container-based & 296 & 75 & 371 \\
  & Spatial-based & 296 & 75 & 371 \\
  & Object-based & 296 & 75 & 371 \\
  & Color-based & 296 & 75 & 371 \\
\midrule
\multirow{4}{*}{Operation}

  & Addition & 320 & 80 & 400 \\
  & Subtraction & 608 & 152 & 760 \\
  & Multiplication & 112 & 28 & 140 \\
  & Division & 144 & 40 & 184 \\
\midrule
\multirow{4}{*}{Quantity Range}

  & 1--5 & 352 & 56 & 408 \\
  & 6--10 & 300 & 100 & 400 \\
  & 11--15 & 264 & 87 & 353 \\
  & 16--20 & 268 & 57 & 325 \\
\midrule
\textbf{Total} & & 1184 & 300 & 1484 \\
\bottomrule
\end{tabular}
\caption{Statistics of E2V-Bench dataset.}
\label{tab:tab15}
\end{table}

\begin{table}[h]
\centering
\scriptsize

\begin{tabular}{llr}
\toprule
\textbf{Category} & \textbf{Sub-category} & \textbf{Count} \\
\midrule
\multirow{4}{*}{Visual Type}
  & Container-based & 57 \\
  & Spatial-based & 241 \\
  & Object-based & 389 \\
  & Color-based & 368 \\
\midrule
\multirow{4}{*}{Operation}
  & Addition & 157 \\
  & Subtraction & 759 \\
  & Multiplication & 106 \\
  & Division & 33 \\
\midrule
\multirow{4}{*}{Quantity Range}
  & 1--5 & 609 \\
  & 6--10 & 276 \\
  & 11--15 & 114 \\
  & 16--20 & 56 \\
\midrule
\textbf{Total} & & 1055 \\
\bottomrule
\end{tabular}
\caption{Statistics of curated training dataset.}
\label{tab:tab17}
\end{table}

\subsection{Comparison of E2V-Bench with Existing Benchmarks}

We present the comparison of \ac{E2V-Bench} with related multimodal benchmarks in Tab.~\ref{tab:tab16}.
\begin{table*}
\centering
\scriptsize

\begin{tabular}{p{2.2cm}p{2.8cm}p{2.2cm}p{3cm}p{3.2cm}}
\toprule
\textbf{Benchmark} & \textbf{\#Entries} & \textbf{Domain} & \textbf{Task Type} & \textbf{Suitability for Arithmetic Education} \\
\midrule
Evaluating Numerical Reasoning in T2I~\citep{kajic2024evaluating} & 1,386 & Numbers, objects & T2I generation & Broad coverage of numerical reasoning, but not designed for early arithmetic education. \\
Math2Visual~\citep{wang-etal-2025-generating-pedagogically} & 1,903 & Math word problems & Template-based visual generation & Targets math word problem education but lacks visual diversity due to its heavy reliance on templates and icon datasets. \\
MathVista~\citep{lu2024mathvista} & 6,141 & Math + vision & Multimodal VQA & Evaluation on reasoning, not visual generation. \\
MATHVision~\citep{wang2024measuring} & 3,040 & General mathematics, competition-level problems & Visual math problem-solving; multimodal reasoning evaluation & Targets middle–high school competition-level math, not early arithmetic education. \\
\midrule
\ac{E2V-Bench} (Ours) & 5,738 (1,484 \ac{VD}s + 4,072 \ac{VD}--visual pairs + 182 real educational visuals) & Arithmetic equations & Equation-to-Visual generation & Explicitly targets early arithmetic education with diverse, non-template-based visual generation. \\
\bottomrule
\end{tabular}
\caption{Comparison of \ac{E2V-Bench} with related multimodal benchmarks.}
\label{tab:tab16}
\end{table*}

\section{Experiments Details}
\subsection{Naive Equation-to-Visual Generation}\label{sec:naive_gen}
\begin{figure}[h!]
    \centering
    \includegraphics[width=0.3\textwidth]{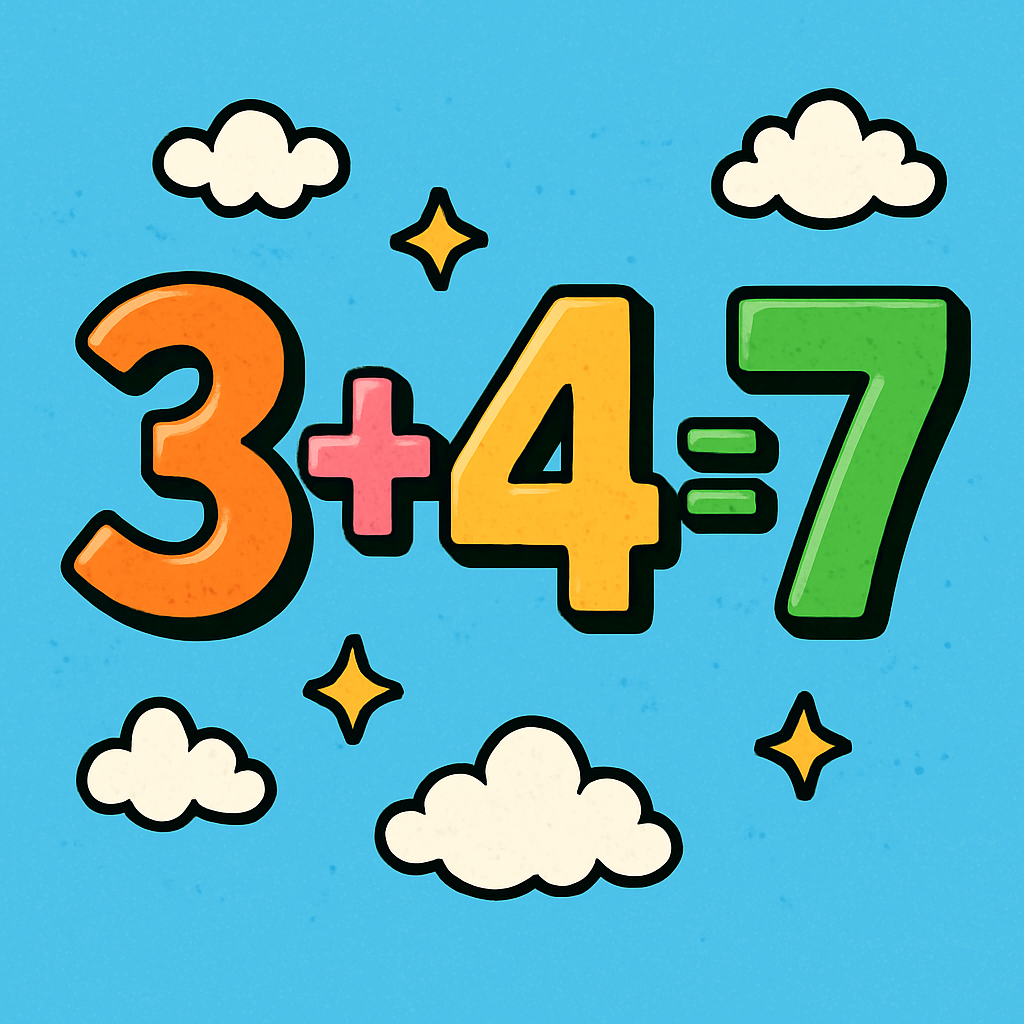}

    \caption{Example of a visual generated by GPT-Image-1 with the prompt: ``Create a cartoon style image to visualize this equation:3+4=7''
    }
    \label{fig:fig2}

\end{figure}

\begin{figure}[h!]
    \centering
    \includegraphics[width=0.3\textwidth]{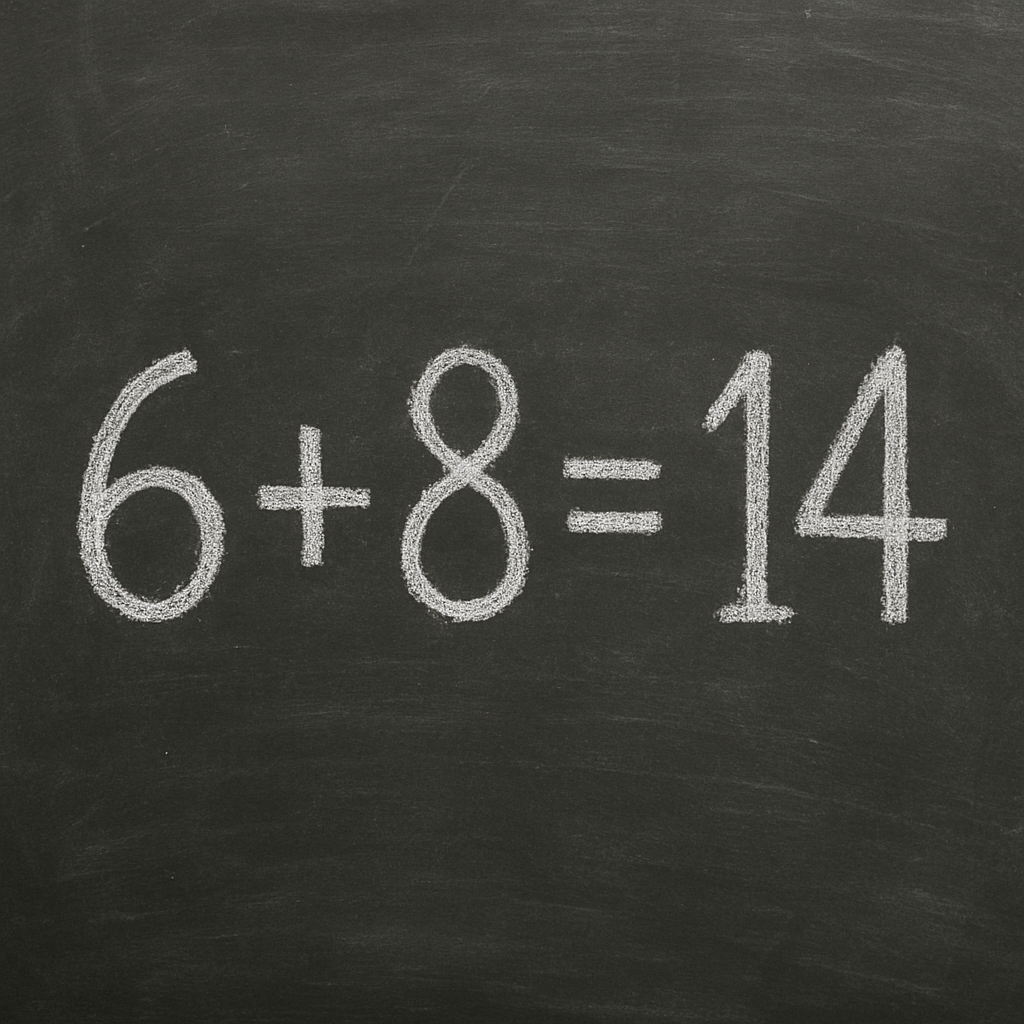}

    \caption{Example of a visual generated by GPT-Image-1 with the prompt: ``Create a realistic style image to visualize this equation:6+8=14''
    }
    \label{fig:fig3}

\end{figure}

We first explored a straightforward approach by directly prompting a \ac{T2I} model to visualize equations. Example outputs are shown in Fig.~\ref{fig:fig2} and Fig.~\ref{fig:fig3}. As illustrated, the generated visuals mainly decorate the equation text with colors or stylistic effects, rather than producing meaningful representations.

\subsection{Automatic Metrics}\label{sec:auto_metric}

\subsubsection{Quantity Accuracy}
For quantity accuracy, we use CountGD~\citep{AminiNaieni24} to automatically count object instances in generated visuals.

\subsubsection{Overall Accuracy}
For overall accuracy, we adopt a rule-based evaluation framework tailored to each visual type, implemented using OpenCV~\citep{pypiOpencvpython} and Scikit-Learn~\citep{scikitlearnScikitlearnMachine}. Overall accuracy is defined at the image level: a generated visual is marked as correct only if it satisfies all task-specific structural and semantic constraints implied by the corresponding equation and visual type.

Across all visual types, we first detect objects using a counting-oriented detection model (CountGD~\citep{AminiNaieni24}), producing object counts and bounding boxes that serve as shared inputs to downstream evaluations. Based on the annotated visual type, we then apply specialized verification rules, as described below.

\paragraph{Object-based visuals.}
For object-based visuals, overall accuracy is equivalent to quantity accuracy. A visual is considered correct if the detected number of each object type exactly matches the ground-truth counts derived from the equation.

\paragraph{Color-based visuals.}
For color-based visuals, we evaluate both count and color assignment. Using object bounding boxes from the counting stage, we extract object crops and estimate dominant colors in HSV space with Gaussian-weighted aggregation. A visual is marked correct only if (1) the number of detected objects matches the ground truth and (2) the multiset of predicted colors matches the expected color distribution.

\paragraph{Spatial-based visuals.}
For spatial-based visuals, we verify whether objects are correctly grouped according to spatial relations (e.g., left/right or separated groups). We apply a dual clustering strategy that combines spatial proximity and geometric similarity. Specifically, we first perform agglomerative hierarchical clustering with single linkage on object center coordinates to capture spatial adjacency, which is effective for separated or chain-like layouts. If this strategy fails to recover the expected grouping, we additionally apply k-means clustering over object bounding box geometry (width and height) to distinguish groups based on size and position. A visual is considered correct only if the predicted group sizes exactly match the ground-truth grouping specified by the equation.

\paragraph{Container-based visuals.}
For container-based visuals, we jointly evaluate object–container assignment and count consistency. Container regions are detected using a grounding-based detector GroundingDINO~\citep{groundingdino}, and objects are assigned to containers based on spatial overlap and relative position.
We use GroundingDINO for container detection because it generalizes better to container-like objects and achieves higher accuracy than CountGD in our preliminary experiments. A visual is marked correct only if each container contains the exact number of objects specified by the equation.

\subsubsection{Metric Validation} \label{sec:metric_validation_de}

To validate the reliability of our automatic metrics, we compare automatic judgments against human annotations on both overall accuracy and quantity accuracy.
Across all validation examples, Cohen's $\kappa$ between human annotations and automatic metrics is 0.96 for both criteria, indicating strong agreement.
To further examine whether this agreement is consistent across different conditions, we report fine-grained agreement rates by quantity range and visual type.
For these subgroup analyses, we report agreement rates rather than Cohen's $\kappa$, since some subsets contain relatively few examples, making subgroup-level $\kappa$ less stable.
As shown in Tables~\ref{tab:metric_validation_quantity} and~\ref{tab:metric_validation_visual_type}, the automatic metrics maintain high agreement with human annotations across all quantity ranges and visual types.

\begin{table}[h]
\centering
\scriptsize
\begin{tabular}{lrrr}
\toprule
\textbf{Quantity Range} & \textbf{N} & \textbf{Overall Agree. (\%)} & \textbf{Quantity Agree. (\%)} \\
\midrule
1--5   & 10 & 90.0  & 90.0  \\
6--10  & 53 & 100.0 & 100.0 \\
11--15 & 75 & 100.0 & 100.0 \\
16--20 & 78 & 100.0 & 100.0 \\
\bottomrule
\end{tabular}
\caption{Metric--human agreement by quantity range. Agreement rates are reported for both overall correctness and quantity correctness.}
\label{tab:metric_validation_quantity}
\end{table}

\begin{table}[h]
\centering
\scriptsize
\begin{tabular}{lrrr}
\toprule
\textbf{Visual Type} & \textbf{N} & \textbf{Overall Agree. (\%)} & \textbf{Quantity Agree. (\%)} \\
\midrule
Color-based     & 54 & 100.0 & 100.0 \\
Container-based & 54 & 100.0 & 100.0 \\
Spatial-based   & 54 & 100.0 & 100.0 \\
Object-based    & 54 & 98.1  & 98.1  \\
\bottomrule
\end{tabular}
\caption{Metric--human agreement by visual type. Agreement rates are reported for both overall correctness and quantity correctness.}
\label{tab:metric_validation_visual_type}
\end{table}

\subsection{Prompts for Generating \ac{VD}}

The prompt for generating object-based \ac{VD}:

\begin{lstlisting}
You are a primary school math teacher. Your task is to create a simple visual description that corresponds to a given arithmetic equation. The visual description should use everyday objects that young children (Grades 1--3) can easily understand.

Each operand in the equation should be represented by a different type of object to clearly distinguish the groups of numbers. Do not use colors or containers as distinguishing features; rely only on object types.

In addition to the word problem, you must also provide:
(1) the total count of each object type mentioned

Here are some examples:
...

Now, for the following equation, generate the outputs following the same style:

Equation: {equation}

Please respond in the following format:
visual_description: <natural language visual description>
object_count: { "<object_name>": <count>, ... }

\end{lstlisting}

The prompt for generating color-based \ac{VD}:

\begin{lstlisting}
You are a primary school math teacher. Your task is to create a simple visual description that corresponds to a given arithmetic equation. The visual description should use everyday objects that young children (Grades 1--3) can easily understand.

Each operand should be represented by identical objects, differentiated only by their colors.

In addition to the word problem, you must also provide:
(1) the total count of the objects, and
(2) a structured description of how quantities are grouped by color.

Here are some examples:
...

Now, for the following equation, generate the outputs following the same style:

Equation: {equation}

Please respond in the following format:
visual_description: <natural language visual description>
object_count: { "<object_name>": <total_count> }
visual_structure: {
  "<object_name>": {
    "total": <total_count>,
    "<color_1>": <count>,
    "<color_2>": <count>
  }
}
\end{lstlisting}

The prompt for generating spatial-based \ac{VD}:

\begin{lstlisting}
You are a primary school math teacher. Your task is to create a simple visual description that corresponds to a given arithmetic equation. The visual description should use everyday objects that young children (Grades 1--3) can easily understand.

Each operand should be represented by placing groups of objects in distinct spatial positions (e.g., left/right, near/far).

In addition to the word problem, you must also provide:
(1) the total count of the objects, and
(2) a structured description of how quantities are grouped spatially.

Here are some examples:
...

Now, for the following equation, generate the outputs following the same style:

Equation: {equation}

Please respond in the following format:
visual_description: <natural language visual description>
object_count: { "<object_name>": <total_count> }
visual_structure: {
  "group1": { "total": <count>, "<object_name>": <count> },
  "group2": { "total": <count>, "<object_name>": <count> }
}
\end{lstlisting}

The prompt for generating container-based \ac{VD}:

\begin{lstlisting}
You are a primary school math teacher. Your task is to create a simple visual description that corresponds to a given arithmetic equation. The visual description should use everyday objects that young children (Grades 1--3) can easily understand.

Each operand should be represented by placing objects into one or more containers.

In addition to the word problem, you must also provide:
(1) the total count of each object type, and
(2) a structured description of how objects are distributed across containers.

Here are some examples:
...

Now, for the following equation, generate the outputs following the same style:

Equation: {equation}

Please respond in the following format:
visual_description: <natural language visual description>
object_count: { "<object_name>": <total_count> }
visual_structure: {
  "<container1_name>": { "total": <count>, "<object_name>": <count> },
  "<container2_name>": { "total": <count>, "<object_name>": <count> }
}
\end{lstlisting}

\subsection{Prompts for Generating Visuals}

The prompt for generating realistic style visuals:
\begin{lstlisting}
Create a realistic style image of: <visual description>
\end{lstlisting}

The prompt for generating cartoon style visuals:
\begin{lstlisting}
Create a cartoon style image of: <visual description>
\end{lstlisting}

\subsection{Synthetic Data Generation and Style Conversion} \label{appendix:syn_data}
\paragraph{Synthetic data generation.}
We constructed a synthetic dataset using a template-based generation pipeline grounded in arithmetic equations.
We exclusively used equations from the \ac{E2V-Bench} \emph{training split} to avoid any data leakage.

Given an equation and its annotated visual type, the pipeline deterministically parses the arithmetic operands and operator, and instantiates a corresponding visual scene using a library of icon assets.
This icon library was collected from flaticon~\citep{flaticon} and consists of 20 high-quality PNG icons representing common objects appearing in \ac{E2V-Bench}.
Each visual is composed of one or more ``groups'', where each group specifies (i) the object type (icon), (ii) the number of instances, and (iii) optional visual attributes such as color or container membership.
A dynamic layout algorithm places these groups onto a fixed-size canvas, automatically selecting grid dimensions, icon size, padding, and inter-group spacing to ensure balanced composition across a wide range of quantities.

In parallel, we generate a paired \ac{VD} using language templates aligned with the underlying arithmetic operation.
For each sample, we also record structured metadata: including object counts, color assignments, container structure, and spatial grouping, which is serialized into the accompanying CSV file.
To increase diversity, each equation is rendered multiple times with randomized icon choices, color assignments, and layout configurations.

\paragraph{Motivation for style conversion.}
The resulting synthetic images are visually clean and icon-centric, which differs substantially from Bagel's native generation distribution.
Directly fine-tuning Bagel on these images risks overfitting to superficial stylistic artifacts rather than improving the model's visual reasoning over quantities and group structure.
To mitigate this mismatch, we apply Bagel's image-to-image generation module to convert the synthetic images into Bagel-aligned visual styles before using them for training.

\paragraph{Style conversion with Bagel.}
We perform style conversion using the original Bagel model in an image-conditioned generation setting.
Given a synthetic icon-based image and its corresponding \ac{VD}, we prompt the model to regenerate the image in a realistic Bagel style while preserving the underlying semantic structure.
Specifically, each input image is passed through Bagel's vision encoder and VAE, and generation is guided by a text prompt of the form:
\emph{“Change this image to realistic style: [visual description]”}.

\paragraph{Filtering and final dataset.}
After style conversion, we filtered the generated images with our automatic metrics.
This process resulted in 3,017 high-quality images, which form the final synthetic extension used in our follow-up fine-tuning experiments.
By separating structural generation from stylistic alignment, this pipeline allows us to inject large-scale, logically controlled supervision while maintaining compatibility with Bagel's visual generation space.

\subsection{Prompt Refinement Details}\label{sec:prompt_refine}
We present the original prompt and the enhanced prompt for layout generation in LMD.

Original prompt:
\begin{lstlisting}
You are an intelligent bounding box generator. I will provide you with a caption for a photo, image, or painting. Your task is to generate the bounding boxes for the objects mentioned in the caption, along with a background prompt describing the scene. The images are of size 512x512. The top-left corner has coordinate [0, 0]. The bottom-right corner has coordinnate [512, 512]. The bounding boxes should not overlap or go beyond the image boundaries. Each bounding box should be in the format of (object name, [top-left x coordinate, top-left y coordinate, box width, box height]) and should not include more than one object. Do not put objects that are already provided in the bounding boxes into the background prompt. Do not include non-existing or excluded objects in the background prompt. Use "A realistic scene" as the background prompt if no background is given in the prompt. If needed, you can make reasonable guesses. Please refer to the example below for the desired format.

Caption: A realistic image of landscape scene depicting a green car parking on the left of a blue truck, with a red air balloon and a bird in the sky
Objects: [('a green car', [21, 281, 211, 159]), ('a blue truck', [269, 283, 209, 160]), ('a red air balloon', [66, 8, 145, 135]), ('a bird', [296, 42, 143, 100])]
Background prompt: A realistic landscape scene
Negative prompt: 
\end{lstlisting}

Enhanced prompt:
\begin{lstlisting}
You are an expert in drawing bounding box for layout-to-image generation.
            When drawing bounding box, you first think about a reasonable scene that can incorporate all objects, describe the scene in text. Then draw the bounding box for each object, output bounding box and the scene prompt.
            
            Note: 
            1. You can draw limited amount of other objects to make the whole image realistic, but the quantity of objects specified in the prompt should be accurate. 
            2. Bounding box should reflect the shape of the object, and the object mentioned in the prompt should be the focus of the image and their bounding box should be BIG for visualization.
            3. If not specified in the prompt, make sure same type of objects are grouping together. If the prompt involve same type of objects in different color, group objects of the same color together.
            4. Please place the bounding boxes in a natural and spatially sensible way: for example, objects should not be floating in the air. If there are too many objects, you can use a top-down view as indicated in the Background prompt. Similarly, if the objects are inside a container, you may also use a top view to make both the container and the objects visible.
            5. Make sure no bounding box exceeds the image boundary.
            
            Example:
            1. Input prompt: There are five balloons floating in the air. A short distance away, there are eight green balloons also floating.
            Output Bounding Box:[('balloon', [8, 62, 95, 100]), ('balloon', [115, 62, 96, 102]), ('balloon', [9, 177, 93, 98]), ('balloon', [118, 176, 96, 101]), ('balloon', [14, 293, 97, 97]), ('balloon', [294, 27, 97, 103]), ('balloon', [403, 24, 100, 107]), ('balloon', [291, 139, 102, 98]), ('balloon', [410, 137, 95, 102]), ('balloon', [285, 244, 107, 104]), ('balloon', [405, 244, 97, 105]), ('balloon', [284, 361, 110, 93]), ('balloon', [407, 357, 100, 96])]
            Output Background Prompt: A realistic scene
            Negative prompt: 
\end{lstlisting}

\subsection{Details of SFT on Flux.1-dev} \label{sec:flux_sft}
We fine-tuned Flux.1-dev (12B parameters) for 10 epochs with 10 repetitions, using a batch size of 5. This batch size balanced GPU memory constraints (single RTX 4090) and batch diversity, which we found to improve convergence stability in preliminary runs. Gradient checkpointing was enabled to reduce memory consumption and allow deeper model tuning without sacrificing input resolution.

For optimization, we used the AdamW\_BF16 optimizer~\citep{loshchilov2019decoupled} with an initial learning rate of 1e-5. Learning rate was decayed with a polynomial scheduler and no warmup, which yielded smoother training dynamics compared to linear decay or cosine schedules. All fine-tuning was conducted with BF16 precision.

To enable parameter-efficient fine-tuning, we adopted LoRA adapters via the Lycoris framework~\citep{yeh2023navigating}, updating only the attention layers instead of full model weights. Images were generated at 1024$\times$1024 resolution to ensure high-quality visuals suitable for educational use cases while remaining computationally feasible. Fine-tuning required approximately 48 hours per run on a single NVIDIA RTX 4090 GPU.

\subsection{Details of SFT on Bagel} \label{sec:bagel_sft}
We fine-tuned BAGEL-MoT (7B parameters) for 10,000 optimizer steps. We used a packed-batch scheme of one sequence per rank targeting 10,240 tokens (flush at 11,520; per-sample cap 10,240) to balance GPU memory limits with sequence diversity observed to aid convergence in preliminary runs. FSDP and non-reentrant activation checkpointing were enabled to reduce memory footprint and permit longer effective contexts without lowering input fidelity.
Images were generated at 1024$\times$1024 resolution.

For optimization, we used AdamW\_BF16 optimizer~\citep{loshchilov2019decoupled} with an initial learning rate of $2 \times 10^{-5}$ ($\beta_{1}=0.9$, $\beta_{2}=0.95$, $\epsilon=1 \times 10^{-15}$) and gradient clipping at $1.0$. The learning rate followed a constant schedule with $2000$ warm-up steps. All fine-tuning was conducted in \texttt{bfloat16} precision (\texttt{bf16}) using \texttt{torch.amp.autocast}.

To limit trainable state, we froze the VAE and enabled only the visual generation pathway (i.e., visual understanding disabled). We fine-tuned from the HuggingFace BAGEL checkpoint\footnote{https://huggingface.co/ByteDance-Seed/BAGEL-7B-MoT}, loading EMA weights only before training. Fine-tuning required approximately 12 hours per run on a 4 NVIDIA GH200 GPUs.

\subsection{Ablation Results}
\label{sec:bagel_ablation}

We further ablate the Bagel adaptation pipeline to disentangle the contribution of its major components, including GPT-Image-1 supervision, synthetic augmentation, style conversion, and RSFT. The full training pipeline is illustrated in Fig.~\ref{fig:fig10}, and the ablation results are reported in Tab.~\ref{tab:bagel_adaptation_ablation}.

The results show that fine-tuning on filtered GPT-Image-1 data provides a modest improvement in overall accuracy, from $5.7\%$ to $6.3\%$, while quantity accuracy remains unchanged. In contrast, using synthetic data alone does not improve performance, yielding $5.0\%$ overall accuracy, and directly combining GPT-Image-1 data with synthetic data without style conversion also provides limited benefit. This suggests that the synthetic visuals are not immediately effective as training data due to their distribution gap from Bagel-generated images.

After applying Bagel style conversion to the synthetic data, performance improves substantially, with quantity accuracy increasing to $15.3\%$ and overall accuracy to $10.3\%$. This indicates that style conversion helps align synthetic examples with Bagel's visual distribution, making the structured synthetic supervision more useful. Finally, adding RSFT further improves overall accuracy to $14.7\%$, suggesting that iterative metric-guided selection helps refine the model toward outputs that better satisfy E2V correctness constraints. Overall, these results indicate that the gains come not merely from adding more data, but from combining high-quality GPT-Image-1 supervision, distribution-aligned synthetic augmentation, and iterative refinement.

\begin{table}[h]
\centering
\scriptsize
\setlength{\tabcolsep}{3pt}
\begin{tabular}{lcccccc}
\toprule
\textbf{Base} & \textbf{GPT} & \textbf{Synthetic} & \textbf{Style Conv.} & \textbf{RSFT} & \textbf{Quant.} & \textbf{Overall} \\
\midrule
Bagel & -- & -- & -- & -- & $8.0 \pm 3.1$ & $5.7 \pm 2.6$ \\
Bagel & $\checkmark$ & -- & -- & -- & $8.0 \pm 3.1$ & $6.3 \pm 2.8$ \\
Bagel & -- & $\checkmark$ & -- & -- & $8.7 \pm 3.2$ & $5.0 \pm 2.5$ \\
Bagel & $\checkmark$ & $\checkmark$ & -- & -- & $8.7 \pm 3.2$ & $5.7 \pm 2.6$ \\
Bagel & $\checkmark$ & $\checkmark$ & $\checkmark$ & -- & $15.3 \pm 4.1$ & $10.3 \pm 3.5$ \\
Bagel & $\checkmark$ & $\checkmark$ & $\checkmark$ & $\checkmark$ & $16.7 \pm 4.3$ & $14.7 \pm 4.0$ \\
\bottomrule
\end{tabular}
\caption{\textbf{Ablation of the Bagel adaptation pipeline on realistic-style E2V-Bench.}
GPT-Image-1 data supervision, Bagel style conversion, and RSFT each contribute to the final improvement. GPT denotes GPT-Image-1 SFT, Synthetic denotes synthetic augmentation, Style Conv. denotes Bagel style conversion, and RSFT denotes rejection-sampling supervised fine-tuning.}
\label{tab:bagel_adaptation_ablation}
\end{table}

\subsection{Cartoon Style Results} \label{sec:cartoon_result}

\begin{figure}[t]
    \centering

    \includegraphics[width=\linewidth]{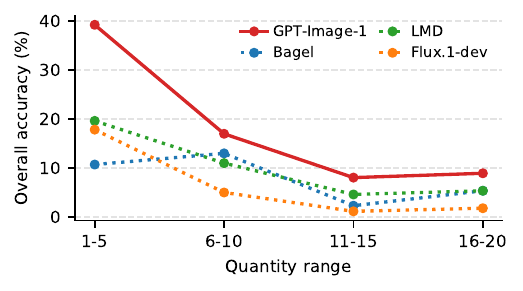}

    \caption{\textbf{Overall accuracy across quantity ranges for cartoon-style visuals.} The corresponding realistic-style results are shown in Fig.~\ref{fig:fig5}.}
    \label{fig:fig7}

\end{figure}

\begin{figure}[t]
    \centering

    \includegraphics[width=\linewidth]{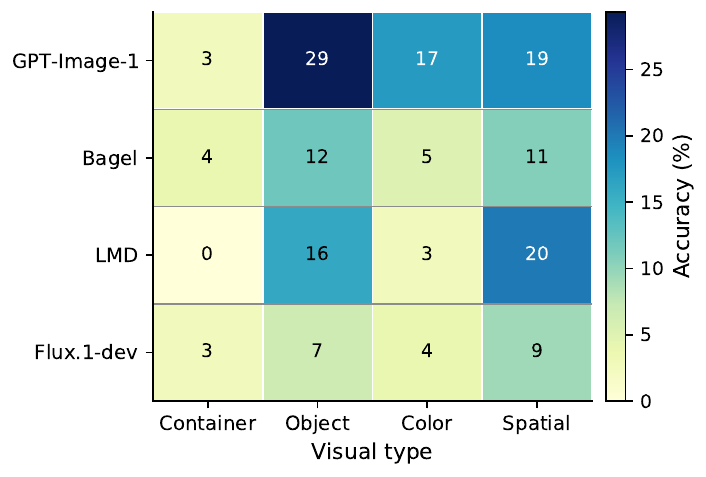}

    \caption{\textbf{Overall accuracy across visual types for cartoon-style visuals.} The corresponding realistic-style results are shown in Fig.~\ref{fig:fig6}.}
    \label{fig:fig8}
    \vspace{-15pt}
\end{figure}

Tab.~\ref{tab:tab1_cartoon} presents the evaluation of T2I models on E2V-Bench for cartoon-style outputs.
Figures~\ref{fig:fig7} and~\ref{fig:fig8} present the overall accuracy of cartoon style visuals generated by different models, evaluated across varying quantity ranges and visual types, respectively.

\begin{table*}
\centering
\scriptsize
\begin{tabular}{p{2.4cm}p{3.5cm}|cc}
\toprule
\textbf{Category} & \textbf{Model} & 
\textbf{Quantity Acc. (\%)} & 
\textbf{Overall Acc. (\%)} \\
\midrule
DSL-based
& DSL-based method (ours)
& $\mathbf{98.3 \pm 1.5}$ & $\mathbf{98.3 \pm 1.5}$ \\
\midrule
Prompt refinement 
& PAE~\citep{mo2024dynamic}          
& $3.0 \pm 1.9$ & $1.0 \pm 1.1$ \\
\midrule
Closed-source   
& Recraft-v3~\citep{recraftv3}   
& $5.7 \pm 2.6$ & $2.7 \pm 1.8$ \\
& GPT-Image-1~\citep{image-1}  
& $22.3 \pm 4.7$ & $17.0 \pm 4.2$ \\
\midrule
Diffusion-based 
& SD-3.5-large~\citep{esser2024scaling} 
& $8.3 \pm 3.1$ & $5.0 \pm 2.5$ \\
& Flux.1-dev~\citep{FLUX1dev}   
& $9.3 \pm 3.3$ & $5.7 \pm 2.6$ \\
\midrule
Layout-to-image 
& Blueprint~\citep{gani2024llm}    
& $1.3 \pm 1.3$ & $1.0 \pm 1.1$ \\
& LMD~\citep{lian2024llmgrounded}          
& $15.3 \pm 4.1$ & $9.7 \pm 3.4$ \\
\midrule
Transformer-based 
& Show-o2~\citep{xie2025show}      
& $4.3 \pm 2.3$ & $3.0 \pm 1.9$ \\
& Bagel~\citep{deng2025bagel}        
& $11.0 \pm 3.6$ & $8.0 \pm 3.1$ \\
\bottomrule
\end{tabular}
\caption{\textbf{Evaluation of T2I models on E2V-Bench for cartoon-style outputs.} We report quantity accuracy and overall accuracy with 95\% confidence interval half-widths. As in the realistic-style setting, most T2I models struggle to preserve quantity and structure. SD-3.5-large denotes Stable Diffusion-3.5-large.}
\label{tab:tab1_cartoon}
\vspace{-5pt}
\end{table*}

\subsection{Human Evaluation Details}\label{sec:human_eval}
We conducted a human evaluation on a manually curated educational dataset of \ac{VD}–visual pairs, as described in Sec.~\ref{sec:task_define}. 
The dataset was sourced from primary-school mathematics textbooks and online educational platforms.

During manual curation, we collected visuals from arithmetic education sections covering the four basic operations that were accompanied by either an \ac{VD} or a symbolic equation.
For visuals that were associated only with equations and lacked an explicit \ac{VD}, we manually authored a descriptive \ac{VD} consistent with the visual content.
Among the 182 collected visuals, 52 had corresponding \ac{VD}s, while the remaining visuals were originally paired only with equations.

\begin{figure*}[t]
    \centering
    \includegraphics[width=0.9\linewidth]{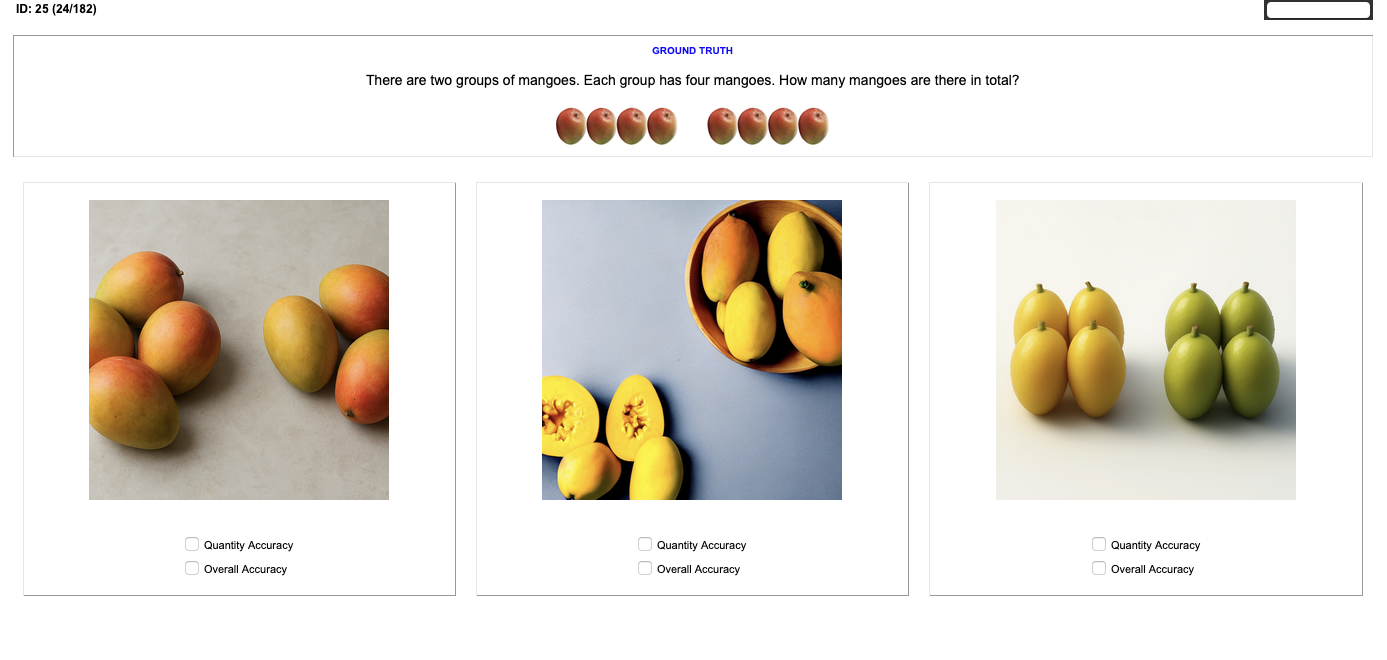}

    \caption{\textbf{Human evaluation interface.} Annotators compared each generated visual against the corresponding ground-truth educational visual and assigned binary judgments for Quantity Accuracy and Overall Accuracy. Clicking the button indicates a correct judgment, while leaving it unclicked indicates an incorrect one. Visuals are shown in random order for each question.}
    \label{fig:fig12}
    \vspace{-15pt}
\end{figure*}

During human evaluation, \ac{VD}s were used as input prompts to the models.
Two researchers independently evaluated a total of 546 generated visuals by comparing them against the ground-truth images and assigning binary judgments for Quantity Accuracy and Overall Accuracy.
A total of three annotation disagreements were identified and resolved through discussion.
A screenshot of the annotation interface used in the human evaluation is provided in Fig.~\ref{fig:fig12}.

\subsection{Example Visuals from \ac{T2I} Models}
\begin{figure*}[t]
    \centering
    \includegraphics[width=0.85\linewidth]{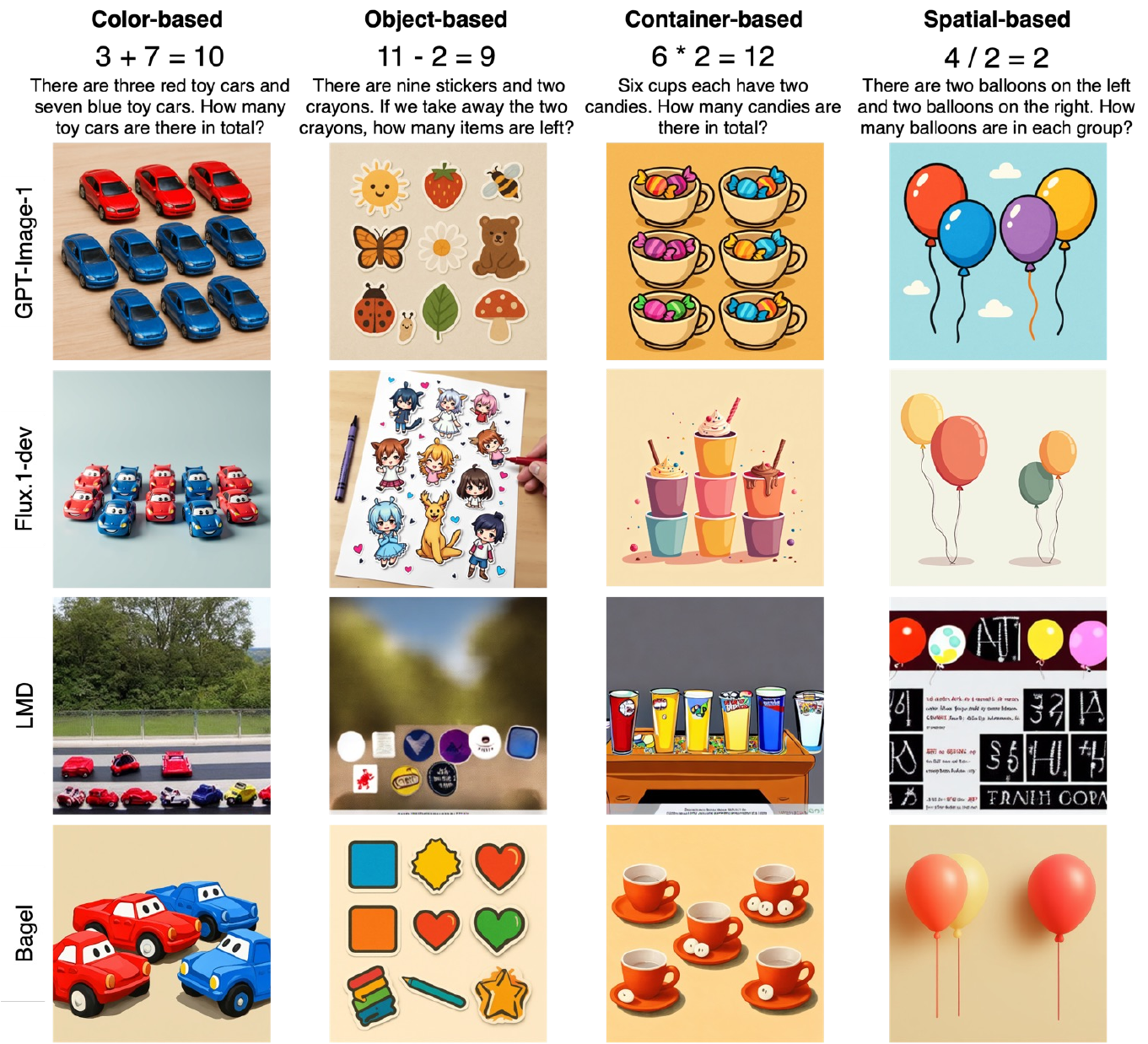}

    \caption{Example visuals from four T2I models on the {E2V-Bench} task. Each column contains one of the four identified structures (Color-based, Object-based, Container-based, Spatial-based), with the input equation and visual description at the top. Each row shows outputs from a different model.
    }
    \label{fig:fig4}
    \vspace{-15pt}
\end{figure*}

We present example visuals generated by \ac{T2I} models on \ac{E2V-Bench} in Fig.~\ref{fig:fig4}.

\subsection{Details of Error Analysis} \label{app:error_analysis}
To better understand the performance drop observed at higher object counts, we conducted a manual error analysis on 208 randomly sampled generated visuals from four representative \ac{T2I} models: Bagel, FLUX.1-dev, GPT-Image-1, and LMD. We sampled 52 error cases per model and analyzed them qualitatively to identify recurring failure patterns.

We used a two-stage coding procedure. First, we performed open coding on 40 randomly selected samples to identify the major error categories emerging from the generated visuals. Based on this process, we established four error types: quantity error, structural error, object mismatch, and other failures. We then applied this codebook to the remaining samples using close coding. Because a single visual could exhibit multiple problems, each sample was allowed to receive more than one error label.

To provide a more fine-grained analysis of quantity errors, we further divided them into minor and major quantity errors. We define a minor quantity error as a mismatch of at most three objects relative to the target visual, and a major quantity error as a mismatch of more than three objects. Structural errors refer to failures in organizing the scene according to the intended visual types, such as missing groupings, incorrect color assignments, or absent containers. Object mismatch captures cases in which the generated object identity is incorrect. Other failures include cases such as no image being generated, blurred outputs, or errors introduced by the automatic metric.

Overall, quantity-related errors dominated. Across all 208 sampled visuals, 33.17\% were labeled as minor quantity error and 57.21\% as major quantity error, while structural error accounted for 16.83\%, object mismatch for 11.06\%, and other failures for 2.88\%. 

\begin{figure}[t]
    \centering

    \includegraphics[width=\linewidth]{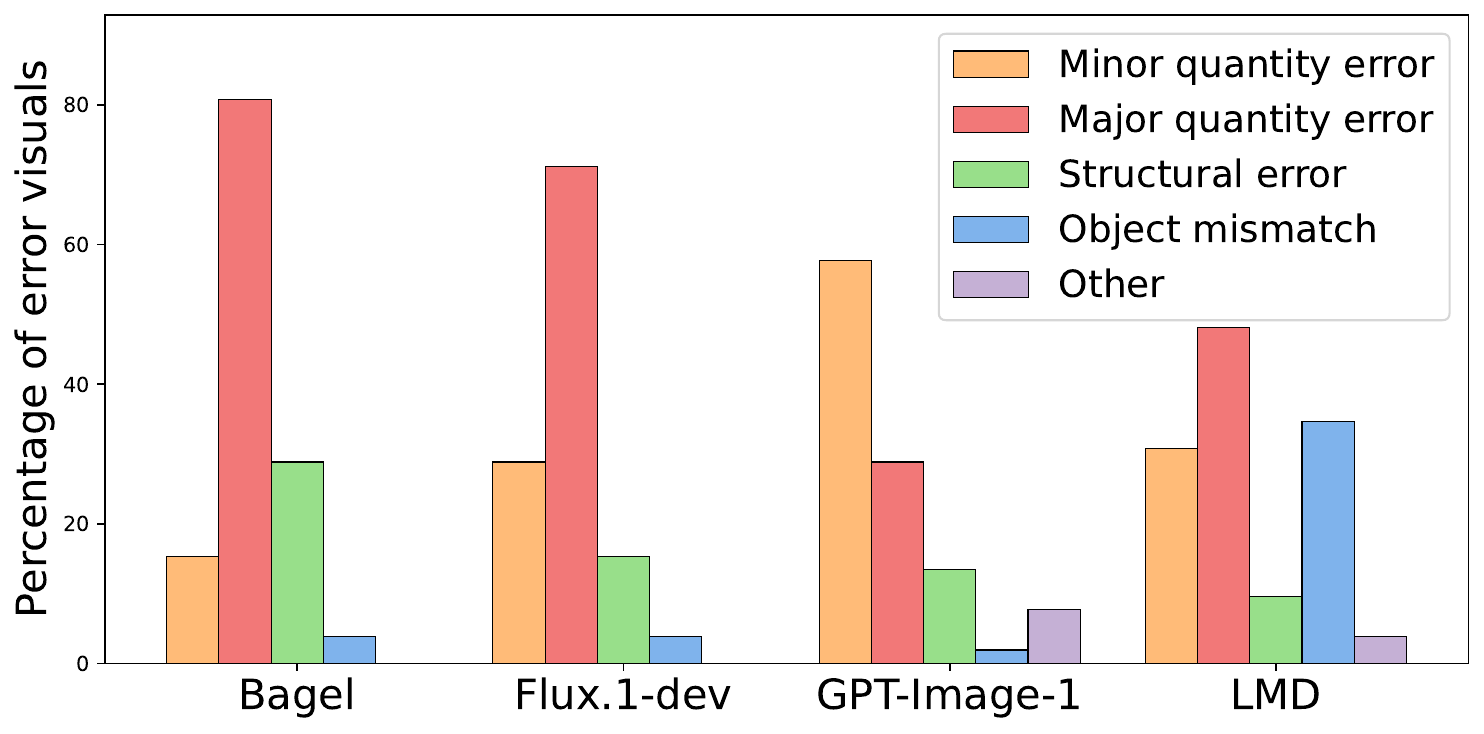}

    \caption{Distribution of manually coded error types across the four representative models. }
    \label{fig:fig13}
    \vspace{-15pt}
\end{figure}

A model-level breakdown further reveals different failure tendencies as shown in Fig.~\ref{fig:fig13}. Bagel and FLUX were dominated by major quantity errors (80.77\% and 71.15\%, respectively), suggesting difficulty in maintaining accurate object counts as scene complexity increases. GPT-Image-1 showed a higher proportion of minor quantity errors (57.69\%), indicating that its failures were more often near-miss counting errors rather than severe deviations. LMD exhibited a comparatively high rate of object mismatch (34.62\%), suggesting that in addition to counting difficulty, it also struggled to preserve the intended object identity. These findings indicate that the sharp performance drop for visuals with more objects is driven primarily by quantity errors, with structural and semantic failures playing a secondary but still important role.

\section{Teacher Interview Study Details}\label{sec:teacher_study}

\subsection{Teacher Demographics}
We recruited ten primary school math teachers through Prolific~\citep{prolificProlificEasily} and paid them 16.63 USD per hour, which is adequate given the participants' country of residence.
We present the participants' demographics in Tab.~\ref{tab:tab5}. All participating teachers use English as the language of instruction.

\begin{table}[h]
    \centering
    \small

    \begin{tabular}{cp{3.5cm}p{1cm}p{1cm}}
     \toprule
      \textbf{PID} & \textbf{Teaching Experience} & \textbf{Age} & \textbf{Gender}\\
      \midrule
       1 & More than 10 years & 46 & Male\\
       2 & 6--10 years        & 36 & Female\\
       3 & 1--2 years         & 28 & Male\\
       4 & More than 10 years & 54 & Female\\
       5 & More than 10 years & 34 & Female\\
       6 & More than 10 years & 51 & Female\\
       7 & More than 10 years & 53 & Male\\
       8 & More than 10 years & 42 & Female\\
       9 & 6--10 years        & 27 & Female\\
      10 & More than 10 years & 53 & Female\\
      \bottomrule       
    \end{tabular}
    \caption{Participants' demographics including teaching experience, age, and gender.}
    \label{tab:tab5}
\end{table}

\subsection{Study Procedure}
Our study obtained ethical approval, and written consent was collected from each participant. The study lasted between 45 minutes and one hour. Participants were first introduced to the background of the study and then completed four sessions, as described below.

In the first session, participants were introduced to the four visual types, accompanied by example visuals for the four arithmetic operations (+, -, $\times$, $\div$). After reviewing each design, they completed a questionnaire to provide feedback.

In the second session, participants were presented with 32 arithmetic problems randomly selected from the \ac{E2V-Bench} test set. They then completed a questionnaire rating the usefulness of these problems for teaching arithmetic skills.

In the third session, participants were shown examples of realistic and cartoon style visuals. They were asked to indicate their preference between the two styles and explain their reasons.

In the fourth session, participants were introduced to the definitions of our proposed evaluation criteria and asked to provide feedback on them.

Finally, after completing all four sessions, participants were asked about their general experiences using visuals to teach arithmetic skills and the types of visuals they typically employ in their classrooms.

\subsection{Additional Results} \label{sec:teacher_result}
\subsubsection{Teachers' Evaluation of Visual Usefulness for Teaching}
We present teachers' ratings of the usefulness of images following our proposed four visual types for teaching math equations in Tab.~\ref{tab:tab6}. 
Scores were collected on a seven-point Likert scale, with seven representing the most positive rating. 
Teachers consistently gave high ratings across all three statements, with averages close to seven (6.7-6.8) and low variance (0.18-0.46). 
These results indicate that, from the teachers' perspective, visuals generated with our proposed visual types are useful for teaching and support students' understanding of math equations. Teachers also indicated a strong willingness to use them in their own instruction.

\begin{table}[h]
    \centering
    \scalebox{0.8}{
    \begin{tabular}{p{5.2cm}cc}
     \toprule
      \textbf{Statement} & \textbf{Average} & \textbf{Variance}\\
      \midrule
       Useful for teaching        & 6.80 & 0.18 \\
       Helpful for student understanding   & 6.70 & 0.23 \\
       Would like to use in teaching      & 6.70 & 0.46 \\
      \bottomrule       
    \end{tabular}
    }
    \caption{Teacher ratings of the usefulness of visuals following the four visual types for teaching arithmetic skills. Scores are on a Likert scale from 1 to 7, where 7 indicates strongly agree.}
    \label{tab:tab6}
\end{table}

\subsubsection{Teachers' Evaluation of Generated Arithmetic Problems for Teaching}
We present teachers' ratings of generated arithmetic problems for teaching in Tab.~\ref{tab:tab7}. 
Scores were collected on a seven-point Likert scale, with seven representing the most positive rating. 
Teachers consistently gave high ratings across all three statements, with averages close to seven (6.7-6.9) and low variance (0.10-0.23). 
These results indicate that, from the teachers' perspective, our generated arithmetic problems are useful for teaching and support students' understanding of math equations. Teachers also indicated a strong willingness to use them in their own instruction.

\begin{table}[h]
    \centering
    \scalebox{0.8}{
    \begin{tabular}{p{5.2cm}cc}
     \toprule
      \textbf{Statement} & \textbf{Average} & \textbf{Variance}\\
      \midrule
       Useful for teaching        & 6.70 & 0.23 \\
       Helpful for student understanding   & 6.70 & 0.23 \\
       Would like to use in teaching      & 6.90 & 0.10 \\
      \bottomrule       
    \end{tabular}
    }
    \caption{Teacher ratings of the usefulness of generated arithmetic problems following the four visual types for teaching arithmetic skills. Scores are on a Likert scale from 1 to 7, where 7 indicates strongly agree.}
    \label{tab:tab7}
\end{table}

\subsubsection{Teachers' Preference for Realistic and Cartoon Style Visuals}
We present teachers' ratings of generated visuals in realistic and cartoon styles for teaching arithmetic skills in Tab.~\ref{tab:tab8}.
Scores were collected on a seven-point Likert scale, with seven representing the most positive rating.
Teachers gave high ratings for both styles, with the realistic style (6.3) rated slightly higher than the cartoon style (6.2), and variance remaining relatively low (0.84–0.90).
Out of ten teachers, four gave higher scores to the realistic style. Six teachers also commented that realistic visuals are useful because students can more easily connect them with familiar objects from daily life (P1–4, P6–7), which helps students better understand the context of the problems.
Three teachers gave higher scores to the cartoon style. Two teachers noted that cartoon visuals are commonly used in their teaching (P5, P10), while two teachers emphasized that the simplified design of cartoon style can reduce student distractions by using plain backgrounds (P10) and identical objects (P7).
These results suggest that both realistic and cartoon visuals are perceived as useful for teaching arithmetic skills, with realistic visuals valued for their connection to students' familiar real-world objects and cartoon visuals appreciated for their clarity and familiarity in educational contexts.

\begin{table}[h]
    \centering
    \scalebox{0.84}{
    \begin{tabular}{p{5cm}cc}
     \toprule
      \textbf{Statement} & \textbf{Average} & \textbf{Variance}\\
      \midrule
       Realistic-style visuals are useful for teaching arithmetic skills.        & 6.30 & 0.90 \\
       Cartoon-style visuals are useful for teaching arithmetic skills.   & 6.20 & 0.84 \\
      \bottomrule       
    \end{tabular}
    }
    \caption{Teacher ratings of the usefulness of generated visuals in realistic and cartoon styles for teaching arithmetic skills. Scores are on a Likert scale from 1 to 7, where 7 indicate most useful.}
    \label{tab:tab8}
\end{table}

\subsubsection{Teacher's Evaluation of Visuals Across Visual Types}
We present teachers' ratings of the usefulness of four visual types for teaching arithmetic operations in Tab.~\ref{tab:tab9}.
Scores were collected on a seven-point Likert scale, with seven representing the most positive rating.

For addition, teachers rated color-based (6.6) and container-based (6.3) visuals as most useful. Seven teachers noted that color-based visuals effectively illustrate the idea of adding new objects into an existing group, while seven teachers emphasized that container-based visuals help clearly separate different groups, making it easier to highlight the part being added.

For subtraction, container-based visuals were rated highest (6.1), followed by spatial-based (5.7). Five teachers mentioned that container-based visuals make the subtraction process easier for students to understand. Three teachers also noted that combining visuals with narrative would be helpful for teaching the subtraction process.

For multiplication and division, container-based visuals received the highest ratings (7.0 for multiplication and 6.9 for division), with spatial-based visuals also considered useful (6.1 for multiplication and 6.4 for division). Seven teachers commented that containers are the most effective way to represent groups in multiplication and division, while four teachers mentioned spatial-based visuals are helpful for showing how items can be grouped or partitioned.

Overall, these results suggest that teachers' evaluations of visual types vary by operation. However, container-based visuals were consistently rated highly across all four operations, with particular strength for multiplication and division.

\begin{table}[h]
    \centering
    \scalebox{0.8}{
    \begin{tabular}{lccccc}
     \toprule
      \textbf{Operation} & \textbf{Color} & \textbf{Object} & \textbf{Spatial} & \textbf{Container} \\
      \midrule
      Addition       & 6.6 & 5.3 & 5.3 & 6.3 \\
      Subtraction    & 5.5 & 5.2 & 5.7 & 6.1 \\
      Multiplication & 5.5 & 4.1 & 6.1 & 7.0 \\
      Division       & 5.6 & 5.1 & 6.4 & 6.9 \\
      \bottomrule       
    \end{tabular}
    }
    \caption{Teacher ratings of the usefulness of four visual types (color-based, object-based, spatial-based, container-based) for teaching different arithmetic operations. Scores are on a Likert scale from 1 to 7, where 7 indicates most useful.}
    \label{tab:tab9}
\end{table}

\subsubsection{Coverage of Visual Types and Teachers' Current Teaching Visuals}
As shown in Tab.~\ref{tab:tab10}, teachers expressed very strong agreement with the statement that the four visual types capture the kinds of visuals they currently use when teaching arithmetic skills in Grades 1–3 (average = 6.90). The low variance (0.10) indicates a high level of consistency among teachers, suggesting that the proposed visual types closely align with existing classroom practices.

All ten teachers reported that they have used visuals similar to our designs following the four visual types. Six teachers mentioned using cartoon-style visuals similar to our designs, while the other four indicated that they have used realistic-style visuals similar to our designs in their classroom teaching.

\begin{table}[h]
    \centering
    \scalebox{0.84}{
    \begin{tabular}{p{5cm}cc}
     \toprule
      \textbf{Statement} & \textbf{Average} & \textbf{Variance}\\
      \midrule
       The four visual types cover the kinds of visuals used to teach arithmetic skills in Grades 1–3.        & 6.90 & 0.10 \\
      \bottomrule       
    \end{tabular}
    }
    \caption{Teachers' ratings of the extent to which the four visual types cover visuals used in teaching arithmetic skills in Grades 1–3. Scores are on a Likert scale from 1 to 7, where 7 indicates strongly agree.}
    \label{tab:tab10}
\end{table}

\section{Ethical Consideration and Applications}
\subsection{Potential Risks}
One potential risk is that generated visuals may be misinterpreted if they do not accurately capture the intended mathematical relationships, potentially leading to confusion among students and educators. To mitigate this risk, we collaborated closely with primary school mathematics teachers to develop a structured design space aligned with pedagogical standards. The automatic metrics we propose further help ensure the correctness of the curated dataset. However, this dataset should not be used directly for educational purposes without the supervision of educators.

\subsection{Terms of Use}

This section outlines the terms and conditions for the use of \ac{E2V-Bench}. By using the code and datasets in this project, users agree to the following terms:

\paragraph{Prohibited Use}

The code and datasets shall not be used for commercial purposes without prior written consent from the authors.

\paragraph{Attribution}
When using or referencing the code and datasets, users must provide proper attribution to the original authors.

\paragraph{No Warranty}
This project is provided as is without any warranties of any kind, either expressed or implied, including but not limited to fitness for a particular purpose. The authors are not responsible for any damage or loss resulting from the use of this project.

\paragraph{Liability}
The authors shall not be held liable for any direct, indirect, incidental, special, exemplary, or consequential damages arising in any way out of the use of the \ac{E2V-Bench}.

\paragraph{Updates and Changes}
The authors reserve the right to make changes to the terms of this license or the \ac{E2V-Bench} itself at any time.

\subsection{Compliance with Artifact Usage and Intended Use Specifications}

\subsubsection{Compliance with Existing Artifact Usage}
In our study, we utilized a range of existing artifacts, such as icons from Flaticon~\citep{flaticon} and visuals from six educational sources~\citep{southsudan_math2_2018,cotton2021oxfordmaths1,mosely2021cambridge2,accessim,mathematicsmonsterMathematicsMonster,fun2dolabs}, to develop our datasets. We rigorously ensured that our usage of these materials was in strict accordance with their intended purposes. Additionally, we employed various computational tools within their prescribed licensing terms, thus adhering to ethical and legal standards.

\subsubsection{Specification of Intended Use for Created Artifacts}
Our research resulted in two primary artifacts: (1) the \ac{E2V-Bench} benchmark and evaluation framework, and (2) a curated training dataset for \ac{T2I} model enhancement.

\paragraph{Dataset License}
We will release \ac{E2V-Bench} and the curated training dataset under the Creative Commons Attribution 4.0 International (CC BY 4.0) license, which permits use, redistribution, adaptation, and commercial use with proper attribution. 
The accompanying evaluation code and scripts will be released under the Apache License 2.0.

\paragraph{E2V-Bench: Benchmark and Evaluation Framework}\mbox{} \par
\textbf{Intended Use:} E2V-Bench is intended for academic research on multimodal learning, text-to-image generation, and educational AI. It supports systematic evaluation of models' ability to generate pedagogically meaningful visual representations from arithmetic equations, and facilitates the development and comparison of model adaptation strategies.

\textbf{Restrictions:} E2V-Bench is designed as a research benchmark and should not be used as a standalone instructional system or deployed directly in classroom settings without additional validation. High-stakes educational or commercial use is discouraged unless supported by further empirical studies and ethical review.

\textbf{Ethical Considerations:} The benchmark design is grounded in analyses of publicly available educational materials and validated through consultations with primary school mathematics teachers. Users are encouraged to preserve the pedagogical intent of the visual types and evaluation criteria when extending or applying the benchmark.

\paragraph{Curated Training Dataset for Model Enhancement}\mbox{} \par
\textbf{Intended Use:} The curated training dataset is intended to support research on improving \ac{T2I} models for equation-to-visual generation. It provides high-quality supervision for studying model adaptation techniques.

\textbf{Restrictions:} The dataset consists of automatically generated and filtered image–text pairs and may still contain inaccuracies or biases inherent to current generative models. It is therefore not recommended for direct instructional use or deployment in educational products without expert oversight.

\textbf{Data Ethics:} All data are synthetically generated or derived from open educational resources and do not contain personally identifiable information. The dataset is released for research purposes, and users are encouraged to adhere to responsible data usage practices consistent with educational and academic norms.

\subsection{Data Collection and Anonymization Procedures}
The benchmark and training datasets do not contain personal data. Arithmetic equations and corresponding word problems were generated by models, and all visuals were produced using \ac{T2I} models.

For components involving human participation, including teacher interviews and evaluations, no personally identifiable information was collected or retained. All feedback was anonymized prior to analysis. In addition, we conducted manual screening to exclude offensive, sensitive, or inappropriate content. These procedures were adopted to ensure ethical data handling, participant privacy, and responsible research practice.

\subsection{Artifact Documentation}

\subsubsection{\ac{E2V-Bench} Benchmark}

\paragraph{Domain Coverage} \ac{E2V-Bench} targets early arithmetic education (Grades 1–3), focusing on equation-to-visual generation for foundational arithmetic concepts.
\paragraph{Operation Coverage} The benchmark covers four basic arithmetic operations: addition, subtraction, multiplication, and division, with quantities restricted to values suitable for primary school instruction.
\paragraph{Visual Types} Each equation is associated with four pedagogically grounded visual types: container-based, object-based, color-based, and spatial-based representations.

\subsubsection{Curated Training Dataset}

\paragraph{Visual Style} The dataset includes realistic style visuals.

\paragraph{Content Scope} All examples are derived from arithmetic equations and corresponding visual descriptions generated to support equation-to-visual learning.

\paragraph{Demographic Representation} The dataset does not encode demographic attributes. Its educational scope reflects arithmetic instruction practices rather than representations of specific populations.

\subsection{Use of AI Assistants in Research}
In our study, AI assistants were used sparingly and in accordance with ACL's Policy on AI Writing Assistance. We utilized ChatGPT and Grammarly for basic paraphrasing and grammar checks, respectively. These tools were applied minimally to ensure the authenticity of our work and to adhere strictly to the regulatory standards set by ACL. Our use of these AI tools was focused, responsible, and aimed at supplementing rather than replacing human input and expertise in our research process.

\subsection{Instructions Given To Participants}
\subsubsection{Disclaimer for Annotators}
Thank you for participating in our evaluation process. Please read the following important points before you begin:

\begin{itemize}
    \item \textbf{Voluntary Participation:} Your participation is completely voluntary. You have the freedom to withdraw from the task at any time without any consequences.
    \item \textbf{Confidentiality:} All data you will be working with is anonymized and does not contain any personal information. Your responses and scores will also be kept confidential.
    \item \textbf{Risk Disclaimer:} This task does not involve any significant risks. It primarily consists of reading and scoring generated visuals.
    \item \textbf{Queries:} If you have any questions or concerns during the task, please feel free to reach out to us.
\end{itemize}

\subsubsection{Instructions for Experiments}
Thank you for participating in our study. This research received ethical approval, and informed consent was obtained from all participants. The study took approximately one hour and consisted of four tasks. Please read the instructions below carefully.

\paragraph{Task 1: Visual Type Evaluation}
You will be presented with visuals representing arithmetic equations across four visual types (container-based, object-based, color-based, and spatial-based) and the four basic arithmetic operations. For each set of visuals, please complete a questionnaire evaluating its usefulness for teaching arithmetic concepts in the classroom.

\paragraph{Task 2: Visual Description Evaluation}
You will review 32 automatically generated visual descriptions corresponding to equations in our benchmark. Please rate whether each word problem is suitable for classroom use and alignment with instructional practice.

\paragraph{Task 3: Visual Style Comparison}
You will evaluate two visual styles: realistic and cartoon, for visualizing arithmetic equations. Please provide feedback on how effective each style may be for educational use, including perceived benefits and potential drawbacks in classroom settings.

\paragraph{Task 4: Evaluation Criteria Feedback}
We will present the definitions of our proposed automatic evaluation metrics for generated visuals. Please comment on whether these criteria adequately capture visual usefulness for teaching arithmetic and provide any suggestions for improvement.

Please answer all questions honestly and feel free to share additional feedback throughout the study. Your responses will help validate our design choices and evaluation framework. Thank you for your time and valuable input.
\subsubsection{Data Consent}
The data you provide during this study will be used solely for academic research purposes. All information will be anonymized and securely stored, and any published or shared data will be aggregated to ensure your privacy. By participating, you agree to the use of your data as described, but you retain the right to withdraw your consent at any time without penalty. If you have any questions about how your data will be used, please feel free to ask the research team.

\begin{figure*}[t]
    \centering
    \includegraphics[width=\linewidth]{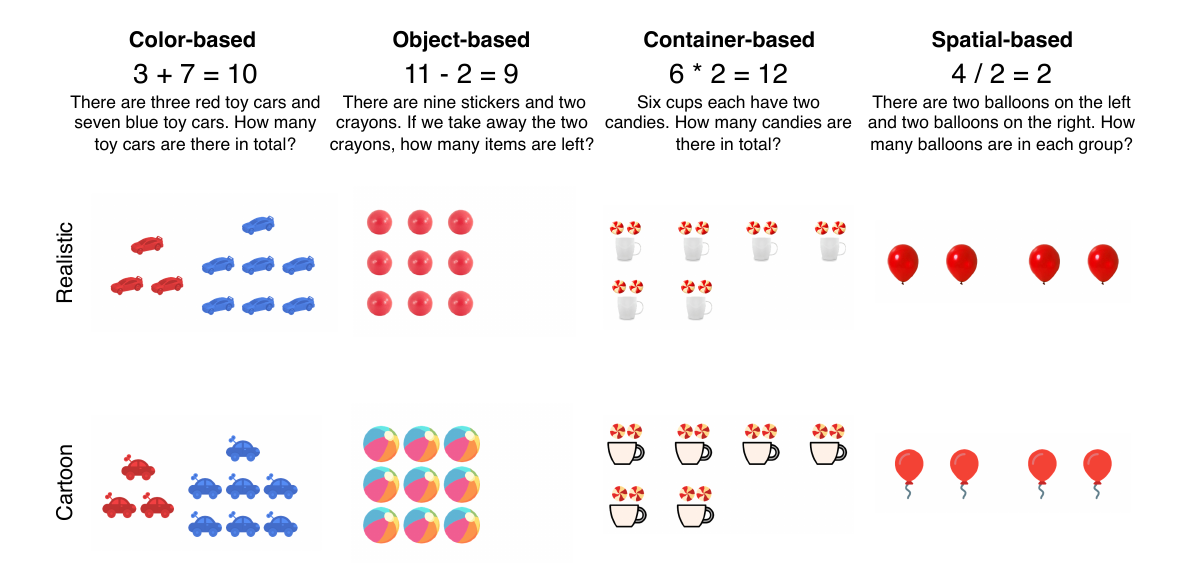}

    \caption{Example visuals generated by the DSL-based pipeline across different visual types and styles.}
    \label{fig:fig14}
    \vspace{-15pt}
\end{figure*}
\section{Details of DSL and DSL-based Visual Generation}
\label{sec:dsl_and_generation}

\subsection{Details of DSL} 
\label{sec:dsl_detail}

To support automatic generation and evaluation, we represent each equation--visual type pair using a simple \ac{DSL}. The \ac{DSL} explicitly encodes the visual type and the grouped object specifications required to represent the operands in the equation:
\vspace{-5pt}
\[
t\{g_i\{\texttt{item}, \texttt{count}, a_t\}\}_{i=1}^{n},
\]
where $t$ denotes the visual type, $g_i$ denotes the $i$-th operand group, \texttt{item} and \texttt{count} specify the object category and quantity within the group, and $a_t$ denotes the type-specific grouping attribute. Concretely, $a_t$ is instantiated as \texttt{color} for color-based visuals, \texttt{position} for spatial-based visuals, and \texttt{container} for container-based visuals. For object-based visuals, the object category itself serves as the grouping cue, so no additional grouping attribute is required.

For example, the equation $3+2=5$ can be represented differently depending on the visual type:
\vspace{-3pt}
\begin{lstlisting}[basicstyle=\ttfamily\small, breaklines=true, columns=fullflexible]
color{
  group1{item: apple, count: 3, color: red}
  group2{item: apple, count: 2, color: blue}
}

object{
  group1{item: apple, count: 3}
  group2{item: banana, count: 2}
}

spatial{
  group1{item: apple, count: 3, position: left}
  group2{item: apple, count: 2, position: right}
}

container{
  group1{item: apple, count: 3, container: basket}
  group2{item: apple, count: 2, container: box}
}
\end{lstlisting}
\vspace{-3pt}

This representation provides a compact semantic specification of the expected visual output. It is used both as a reference representation for automatic evaluation and as the input to our DSL-based visual generation pipeline.

\subsection{DSL-based Visual Generation} 
\label{sec:dsl_gen_detail}

We further implement a DSL-based visual generation pipeline as a structured reference method. Unlike end-to-end \ac{T2I} models, this method does not generate images directly from natural-language prompts. Instead, it renders visuals from the DSL using manually specified layout rules and icon assets. Therefore, it is not intended as a directly comparable \ac{T2I} baseline, but rather as a reference method for assessing whether the visual semantics defined in our benchmark are well specified and renderable.

Given a DSL instance, the renderer first identifies the target visual type $t$ and parses the operand groups $g_i$. It then selects corresponding icon assets based on the \texttt{item} field and places the required number of icons according to \texttt{count} and the grouping attribute $a_t$. For color-based visuals, objects from different groups are rendered with different colors according to the \texttt{color} field. For spatial-based visuals, groups are placed in distinct spatial regions, such as left and right. For container-based visuals, objects are placed inside the specified containers. For object-based visuals, different object categories are used to distinguish operand groups.

To support stylistic variation, we manually collected 176 icons spanning both realistic and cartoon styles. The same DSL can therefore be rendered into different visual styles while preserving the underlying quantities and grouping structure. Example visuals generated by our DSL-based pipeline are shown in Fig.~\ref{fig:fig14}.

\end{document}